\PassOptionsToPackage{unicode}{hyperref}
\PassOptionsToPackage{hyphens}{url}
\documentclass[]{article}
\usepackage{amsmath,amssymb}
\usepackage{lmodern}
\usepackage{tabularx}
\usepackage{booktabs}
\usepackage{float}
\usepackage{iftex}
\usepackage{cite}
\ifPDFTeX
  \usepackage[T1]{fontenc}
  \usepackage[utf8]{inputenc}
  \usepackage{textcomp}
\else
  \usepackage{unicode-math}
  \defaultfontfeatures{Scale=MatchLowercase}
  \defaultfontfeatures[\rmfamily]{Ligatures=TeX,Scale=1}
\fi
\IfFileExists{upquote.sty}{\usepackage{upquote}}{}
\IfFileExists{microtype.sty}{%
  \usepackage[]{microtype}
  \UseMicrotypeSet[protrusion]{basicmath}
}{}
\makeatletter
\@ifundefined{KOMAClassName}{%
  \IfFileExists{parskip.sty}{%
    \usepackage{parskip}
  }{%
    \setlength{\parindent}{0pt}
    \setlength{\parskip}{6pt plus 2pt minus 1pt}}
}{%
  \KOMAoptions{parskip=half}}
\makeatother
\usepackage{xcolor}
\usepackage{longtable,booktabs,array}
\usepackage{calc}
\usepackage{etoolbox}
\makeatletter
\patchcmd\longtable{\par}{\if@noskipsec\mbox{}\fi\par}{}{}
\makeatother
\IfFileExists{footnotehyper.sty}{\usepackage{footnotehyper}}{\usepackage{footnote}}
\makesavenoteenv{longtable}
\usepackage{graphicx}
\makeatletter
\def\maxwidth{\ifdim\Gin@nat@width>\linewidth\linewidth\else\Gin@nat@width\fi}
\def\maxheight{\ifdim\Gin@nat@height>\textheight\textheight\else\Gin@nat@height\fi}
\makeatother
\setkeys{Gin}{width=\maxwidth,height=\maxheight,keepaspectratio}
\makeatletter
\def\fps@figure{htbp}
\makeatother
\ifLuaTeX
  \usepackage{selnolig}
\fi
\IfFileExists{bookmark.sty}{\usepackage{bookmark}}{\usepackage{hyperref}}
\IfFileExists{xurl.sty}{\usepackage{xurl}}{}
\hypersetup{
  pdftitle={IELTS Writing Revision Platform with Automated Essay Scoring and Adaptive Feedback},
  hidelinks,
  pdfcreator={LaTeX via pandoc}}

\usepackage{titling}
\usepackage[top=1in, bottom=1in, left=1in, right=1in]{geometry}

\pretitle{\begin{center}\LARGE\bfseries}
\posttitle{\par\end{center}\vskip 0.5em}
\preauthor{\begin{center}\large}
\postauthor{\end{center}\vskip 1em}
\predate{\begin{center}\large}
\postdate{\end{center}}

\title{IELTS Writing Revision Platform with Automated Essay Scoring and Adaptive Feedback}
\author{Titas Ramancauskas, Kotryna Ramancauske}
\date{}

\begin{document}
\maketitle
\vspace{-2em}

\noindent\textbf{Abstract}---This paper presents the design, development, and evaluation of a proposed revision platform assisting candidates for the International English Language Testing System (IELTS) writing exam. Traditional IELTS preparation methods lack personalised feedback, catered to the IELTS writing rubric. To address these shortcomings, the platform features an attractive user interface (UI), an Automated Essay Scoring system (AES), and targeted feedback tailored to candidates and the IELTS writing rubric. The platform architecture separates conversational guidance from a dedicated writing interface to reduce cognitive load and simulate exam conditions. Through iterative, Design-Based Research (DBR) cycles, the study progressed from rule-based to transformer-based with a regression head scoring, mounted with adaptive feedback. 

Early cycles (2-3) revealed fundamental limitations of rule-based approaches: mid-band compression, low accuracy, and negative $R^2$ values. DBR Cycle 4 implemented a DistilBERT transformer model with a regression head, yielding substantial improvements with MAE of 0.66 and positive $R^2$. This enabled Cycle 5's adaptive feedback implementation, which demonstrated statistically significant score improvements (mean +0.060 bands, p = 0.011, Cohen's d = 0.504), though effectiveness varied by revision strategy. Findings suggest automated feedback functions are most suited as a supplement to human instruction, with conservative surface-level corrections proving more reliable than aggressive structural interventions for IELTS preparation contexts. Challenges remain in assessing higher-band essays, and future work should incorporate longitudinal studies with real IELTS candidates and validation from official examiners.

\vspace{1em}
\hypertarget{i.-introduction}{%
\section{I. Introduction}\label{i.-introduction}}

\textsc{T}he IELTS is known as the benchmark for English competence for academic and
professional purposes, with over 3.5 million tests administered annually
{\cite{sehrish2024problems}}.

The IELTS exams are prerequisites for academic and professional
opportunities in English-speaking countries, recognised by 11,000
organisations across 140 countries. They impact immigration, study, and
work chances globally. Millions of students depend on their IELTS scores
for life-changing opportunities, especially for higher education, where
a minimum score of 6.5 is often required for admission {\cite{arefsadr2023voices}}.

The IELTS writing exam evaluates task achievement, coherence and
cohesion, lexical analysis, and grammatical accuracy {\cite{lodhi2024identification}}. Task 1
requires analysis of visual data, while Task 2 involves constructing a
piece of structured writing. Despite its importance, the writing portion
consistently receives the lowest mean score across all candidate
backgrounds. Shortcomings amongst candidates include question
misinterpretation, insufficient elaboration, and weak argumentation.
Historically consistent underperformance spotlights the need for
targeted preparation tooling.

\hypertarget{ii.-literature-review}{%
\section{II. Literature Review}\label{ii.-literature-review}}

\hypertarget{a.-the-need-for-adaptive-ietls-writing-preparation}{%
\subsection{A. The Need for Adaptive IETLS Writing
Preparation}\label{a.-the-need-for-adaptive-ietls-writing-preparation}}

The writing component of the IELTS assesses four criteria: Task
Achievement (TA), Coherence and Cohesion (CC), Lexical Resource (LR),
and Grammatical Range and Accuracy (GRA); yet it consistently yields the
lowest average scores across test-taker cohorts {\cite{lodhi2024identification}}. Common
challenges include question misinterpretation, limited vocabulary,
recurring grammatical errors, and unfamiliarity with academic writing
conventions. Traditional classroom preparation provides generalised
instructions and delayed, instructor-dependent feedback that lacks
personalisation, leaving learners unable to remediate individual
weaknesses in real-time {\cite{lodhi2024identification}}.

AI writing tools such as Grammarly have gained popularity for providing
instant feedback on grammar, vocabulary, and structure, supplementing
traditional teaching. Empirical studies show that using AI writing tools
leads to measurable performance improvements. Table I tabulates IELTS
candidates' scores across all categories, which improved by integrating
these writing tools into their workflow. This had a strong impact on the
GRA criteria, with an average score increase of 1.2 points, resulting in
an overall band score improvement of about one full band {\cite{saira2025ai}}.

\begin{table}[H]
\centering
\caption{Pre-and Post-Writing Assessment Scores (Mean $\pm$ SD)~\cite{saira2025ai}}
\includegraphics[width=1\linewidth]{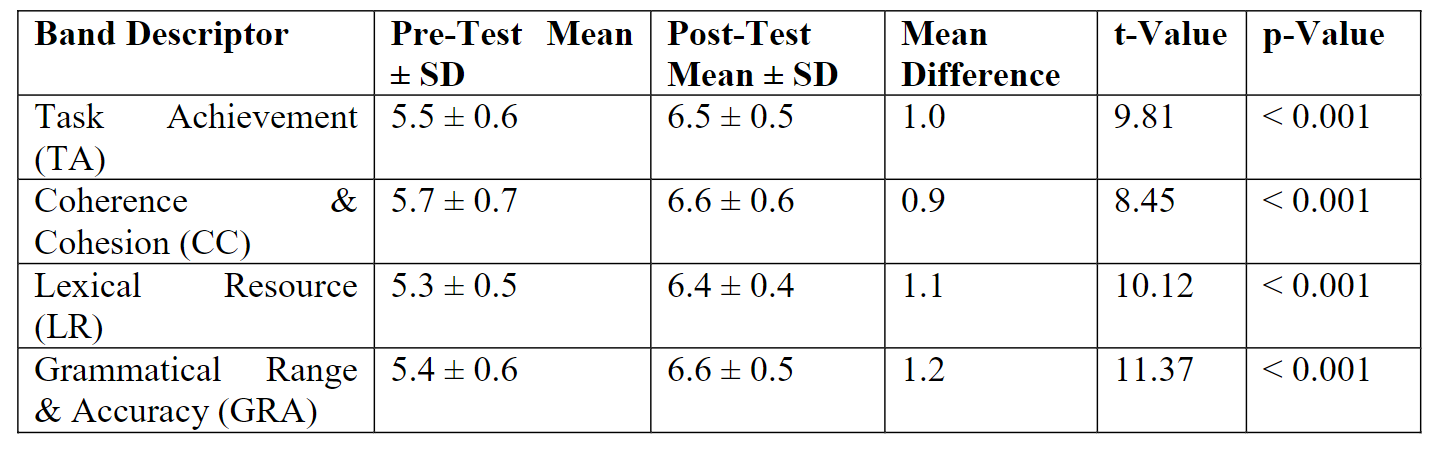}
\label{tab:assessment_scores}
\end{table}

Studies verify AI tools\textquotesingle{} capabilities to increase IELTS
writing scores by about one band, with TA improving from 5.5\,±\,0.66 to
$6.5 \pm 0.51$ and GRA showing the highest gain (mean $\Delta = 1.2$).
However, overreliance on AI may hinder learning towards fundamental
writing skills. Table 2 voices a survey of 60 IELTS candidates found
that while they rated grammar correction (mean\,=\,4.6) and ease of use
(mean\,=\,4.7) highly, they still preferred combining AI feedback with
instructor guidance (mean\,=\,4.5) {\cite{saira2025ai}}.

\begin{table}[H]
\centering
\caption{Students' Perceptions of AI Tools (n=60)~\cite{saira2025ai}}
\includegraphics[width=1\linewidth]{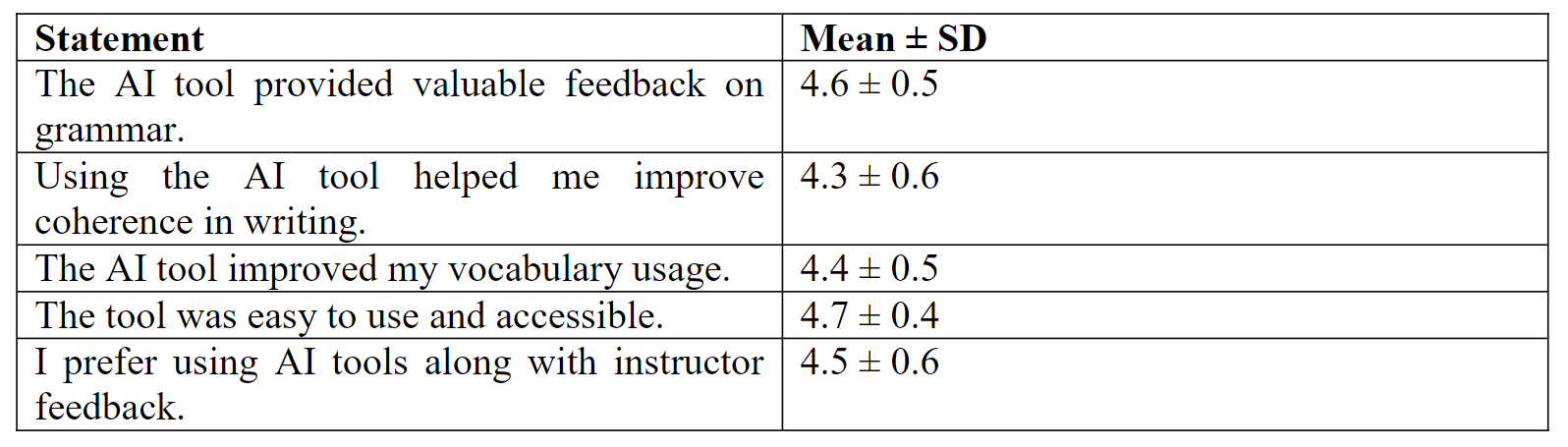}
\label{tab:students_perceptions}
\end{table}

Recent advancements in adaptive educational technologies aim to address
gaps by delivering personalised, calculated feedback aligned with the
IELTS rubrics. Adaptive teaching systems analyse learners' strengths and
weaknesses and adjust feedback accordingly, offering 24/7 support
tailored to individual cognitive profiles {\cite{reyes2024exploring}}. For instance,
Schipper validated that embedding adaptive feedback loops into
linguistic lessons substantially augmented engagement and learning
outcomes compared to static instruction {\cite{schipper2023dealing}}. Yet, few platforms
incorporate adaptive feedback algorithms within systems supporting
realistic writing interfaces. This gap motivates the development of a
platform that provides adaptive feedback coordinated for IELTS writing
practice.

\hypertarget{b.-linguistic-tools-and-ai-support-in-writing}{%
\subsection{B. Linguistic Tools and AI Support in
Writing}\label{b.-linguistic-tools-and-ai-support-in-writing}}

Linguistic learning platforms have emerged to support ESL (English as a
Second Language) learners in areas such as pronunciation, vocabulary,
and grammar, aspects often overlooked in traditional ESL environments.
Research conducted by Ibrahim concluded that ESL learners were hesitant
to use tools like dictionaries, which they perceived as complex,
highlighting the need for innovative solutions {\cite{ibrahim2023artificial}}.

Notably, ELSA Speak and Duolingo are prominent tools that include
immediate feedback and gamification elements to boost learner
engagement. However, despite their advantages, both platforms are
limited in handling complex writing tasks such as IELTS Task 2.

ELSA Speak is a successful language learning platform because users can
concentrate on meaningful, relevant materials tailored to their
profiles. It offers targeted, personalised exercises that focus on both
segmental aspects (individual sounds) and suprasegmental features
(intonation and stress) {\cite{gusrianto2025exploring}}. The platform performs real-time speech
analytics to enhance fluency and intonation through interactive chatbot
conversations. Its success stems from personalisation, a low-stress
learning environment, and adherence to repetition and adaptive
scaffolding, which are fundamental to language learning. Furthermore,
ELSA Speak's chatbot avoids overwhelming learners by providing
understandable outputs and fostering a healthy, minimal-stress learning
environment without cognitive overload {\cite{khoi2024investigating}}. However, ELSA mainly
concentrates on speaking skills and neglects other areas, such as
writing, where IELTS candidates often underperform.

Another prominent linguistic platform is Duolingo, a globally recognised
tool fusing gamification and mobile-assisted learning, supplying a rich
collection of accessible language courses {\cite{shortt2023gamification}}. Research performed
by Laksanasut observed Duolingo's embodiment of gamification aspects
deep-rooted in Self-Determination Theory (SDT), which contributed to
high motivation, minimised anxiety, and high user retention through
daily goals, a reward system, and progress tracking {\cite{laksanasut2025gamification}}.
Duolingo\textquotesingle s SDT gamification approach increases user
persistence; however, its lack of depth disqualifies Duolingo as an
adequate tool for achieving the academic writing and argumentative
complexity required by the IELTS writing tasks.

Both tools lack the depth of contextualised feedback and targeted IELTS
writing support necessary to prepare candidates. However,
ELSA\textquotesingle s personalised AI feedback and exercise generation,
paired with Duolingo's gamification mechanics, model the skeleton of a
holistic, writing-oriented platform.

\hypertarget{c.-conversational-agents-in-education}{%
\subsection{C. Conversational Agents in
Education}\label{c.-conversational-agents-in-education}}

Within educational environments, conversational agents and chatbot
systems have emerged as a favoured preparation technique, especially for
examinations such as the IELTS. Generalised preparation is common in
traditional classrooms, neglecting diverse learning paces, especially
for IELTS candidates who may not identify their linguistic weaknesses
due to inflexibility. Adaptive chatbots, utilising natural language
processing (NLP), provide continuous access to targeted feedback,
thereby addressing challenges of generalisation {\cite{huang2022chatbots}}. Chatbot
technologies, originating with ELIZA in 1964, have evolved to simulate
interactive, meaningful communication, enhancing learner engagement
through dialogue practice and instant feedback.

Research spearheaded by Lin into the architecture of chatbot systems, as
exemplified by the JULIE chatbot in construction management, illustrates
the sophistication possible through entity matching and integration with
structured databases {\cite{lin2023prototyping}}. By processing both structured and
unstructured data from formats such as spreadsheets, documents, and
webpages, it enables the answering of broad user queries while still
providing precise, contextual answers {\cite{lin2023prototyping}}.

Chatbot architectures like JULIE can be integrated into linguistic
contexts such as IELTS to boost candidates\textquotesingle{} confidence
by supporting structured question-answering and personalised responses
to the exam.

\hypertarget{d.-nlp-tools-in-ielts-preparation}{%
\subsection{D. NLP Tools In IELTS
Preparation}\label{d.-nlp-tools-in-ielts-preparation}}

A study by Muhammad demonstrated Google Dialogflow\textquotesingle s
excellent consistency in constructing chatbots in a controlled environment with high accuracy, as evidenced by perfect performance in structured domains, such as hotel check-ins, as shown in Fig. 1, scoring 100\% across all typical procedures, such as passport handling, checking in, thank you, and hello messages {\cite{muhammad2020developing}}.

Studies by Sung and Kang illustrated Dialogflow\textquotesingle s potential as a chatbot
system in an English as a Foreign Language (EFL) setting. Dialogflow balances linguistic complexity with predictable responses, enabling controlled language generation that minimised hallucinations and maintains contextual relevance, which addresses fundamental challenges in language education that traditional classrooms cannot replicate {\cite{sung2024developing}}.

However, while Abdelmoiz found DialogFlow’s metrics impressive in controlled environments {\cite{abdelmoiz2024developing}}, the system’s reliance on structured, predictable patterns makes it poorly suited to the IELTS exam, which requires adaptive, context-aware feedback.

An experiment at Thammasat University assessed writing prep methods: ChatGPT supplemented vs. traditional, led by Bai. Fig. 2 displays that
Pre-test scores were 5.17 (experimental) and 5.22 (control). After prep,
experimental scores rose to 5.8025, control to 5.3. The score difference
is credited to ChatGPT\textquotesingle s influence, also hinting at how
ineffective traditional methods were in comparison {\cite{bai2023application}}.

\begin{table}[H]
\centering
\caption{Pre-and Post-Writing Assessment Scores (Mean $\pm$ SD)~\cite{bai2023application}}
\includegraphics[width=1\linewidth]{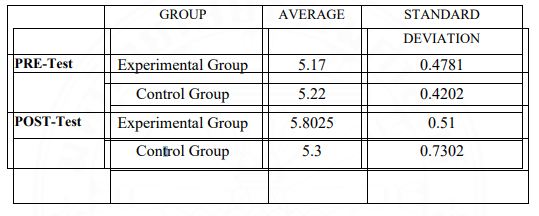}
\label{tab:pre_post_scores}
\end{table}

Bai's study qualitatively dissected the factors contributing to the
superior quality of writing crafted by the experimental group. Core
reasons identified by Bai included the momentous expansion of students'
vocabulary, with the introduction of terms like `celebration',
`symbolise', `atmosphere', and `advertisement'. Additionally,
perfections in writing format and structure directly reflected the
increase in writing scores {\cite{bai2023application}}.

While Bai\textquotesingle s study demonstrated solid results regarding ChatGPT\textquotesingle s integration into linguistic writing preparation, Moran voiced underlying concerns of students utilising such tools, arguing that over-reliance on Generative AI risks inhibiting the development of an authentic writing voice, leaving students “insecure and alienated about their verbal abilities”
{\cite{moran2023chatgpt}}.

The popularity of ChatGPT in education is growing, but relying solely on
it for IELTS prep hinders candidates\textquotesingle{} abilities since
models can\textquotesingle t replicate the human communication skills
required in IELTS testing.

\hypertarget{e.-designing-interfaces-for-feedback-and-learning}{%
\subsection{E. Designing Interfaces for Feedback and
Learning}\label{e.-designing-interfaces-for-feedback-and-learning}}

The user interface and system layout impact how learners interpret and
utilise the feedback delivered. We compare the platform designs of a
website versus a web application to determine which one better supports
chatbots, automated scoring, and adaptive feedback. The stereotypical
website is well-suited to static and semi-interactive resources, for
instance, ASR-assisted materials, but these lack the stateful,
interactive flows necessary in orchestrating conversation, ongoing
scoring, and adaptive feedback {\cite{kasemsarn2023information}}.

To accomplish this functionality, implementing persistent state and
event-driven pipelines is necessary, which surpasses the capabilities of
a static website ecosystem. The web application architecture is more
supportive of stateful interactions without requiring any installs and
scales, empowering users to execute tasks and retrieve feedback
{\cite{vallabhaneni2024secured}}.

The platform enables easy incorporation of features such as chatbots, submission grading, and adaptive feedback loops tailored to the IELTS. Ultimately, a web application was chosen, with a UI designed to aid ESL learners in regulating cognitive load, maximising engagement, and focusing on learning outcomes. 

Futile UIs encourage strenuous cognitive load, which in turn degrades timed-task performance, through inadequate, unclear navigation and task framing, which are imperative to IELTS candidates aiming for high writing standards. 

Due to human biological nature, working memory is limited, carefully
crafted UIs strictly avoid splitting attention and eliminating ambiguity
to optimise cognitive abilities for task performance {\cite{fang2025effects}}.

Specifically targeting cognitive design principles, we prioritised
visual hierarchy (typography, spacing, grouping) to streamline attention
and task flow. Crafting devoted wireframes validated that the UI is
functional, engaging, and reduces task complications, and maximises
candidates' problem-solving abilities {\cite{mohamed2025effect}}.

While chatbots can foster engagement with social interaction, they are
less applicable to the IETLS writing exercise, which requires
transparency, structure, and concentration.

Complete reliance on a chatbot interface risks increasing cognitive load
and losing engagement. This informed the decision to transcend a pure
chatbot interface, dedicating a UI solely to the writing exercise while
being mindful of visual hierarchy, engagement, and cognitive load to aid
candidates' examination success.

\hypertarget{iii.-proposed-framework}{%
\section{III. Proposed Framework}\label{iii.-proposed-framework}}

\hypertarget{a.-introduction-to-the-proposed-framework-writing-preparation}{%
\subsection{A. Introduction to the Proposed Framework }\label{a.-introduction-to-the-proposed-framework-writing-preparation}}

\begin{figure}[H]
\centering
\includegraphics[width=1\linewidth]{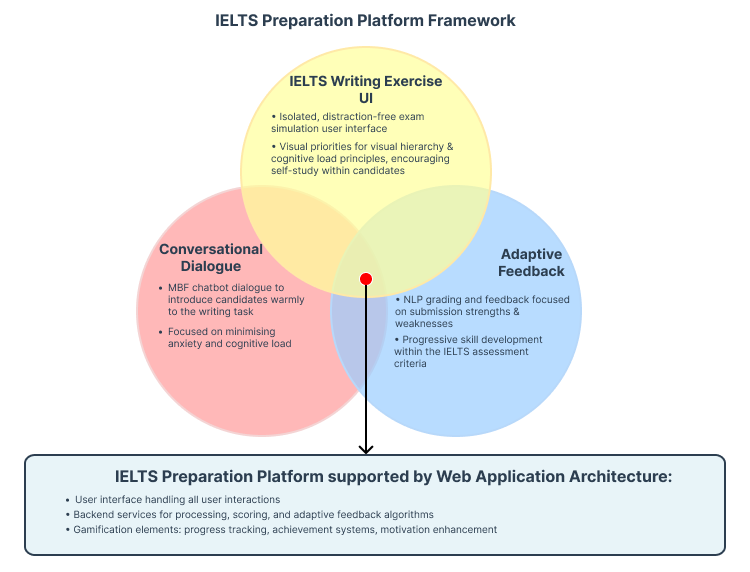}
\caption{IELTS Preparation Platform Framework.}
\label{fig:platform}
\end{figure}

Fig. 1 illustrates the proposed IELTS Preparation platform framework,
including dialogue systems, adpative feedback mechanisms, and a
dedicated IELTS writing exercise screen, within a web application
architecture. The framework domains consists of: (1)
chatbot conversational dialogue system, (2) adaptive feedback
mechanisms based on the IELTS rubric, and (3) writing
exercise interface. The subsequent sections describe the domains, their
intersections, and their impact on candidates.

\begin{enumerate}
\def\labelenumi{\arabic{enumi}.}
\item
  \emph{Conversational Dialogue:} The conversation dialogue domain is
  built using Microsoft Bot Framework, adopting a scaffold comprising of a welcoming
  interaction, prompting users to provide details such as name and age
  to establish familiarity. Users are then presented with a
  preview card of an IELTS writing task, and the ability to select which
  portion they would like to begin with first (introduction, body, or
  conclusion). They are then guided into the main Writing Task UI,
  focusing on their chosen part. Each submission is graded, providing
  adaptive feedback, an estimated grade, and actionable suggestions. By
  decomposing the IELTS writing task into smaller, manageable parts, the
  chatbot reduces cognitive load, alleviates anxiety, and fosters a
  supportive learning environment.
\item
  \emph{Adaptive Feedback:} The adaptive feedback mechanism focuses on
  providing precise evaluations of candidate submissions by comparing
  them against the IELTS rubric, clearly pinpointing strengths and
  weaknesses. Behind the scenes, the mechanism operates using NLP tools
  including BERT, SpaCy, and NLTK, which identify grammatical errors, measure
  vocabulary difficulty, and evaluate coherence. Founding a continuous,
  guided learning environment that adjusts to the candidate's current
  skill level, which reduces cognitive load and helps candidates
  progressively develop their English writing abilities.
\item
  \emph{IELTS Writing Exercise UI:} The IELTS Writing Exercise UI was
  created to help candidates focus on their writing, reducing distractions and
  cognitive load. ESL candidates benefit from a distraction-free
  practice environment. To support this, the platform provides clear
  instructions, a reference image, and a large writing space
  to maximise focus on the task. All features, including submissions,
  progress tracking, grading, and adaptive feedback, are accessible
  through a dismissible sidebar. Adhering to cognitive load and
  visual hierarchy principles, guiding the user's attention to key
  objectives. This core interface connects user interactions, adaptive feedback, and automated grading, while
  promoting self-study strategies essential for IELTS success.
\item
  \emph{Platform Architecture:} The IELTS preparation platform features
  a web application architecture that uses a React frontend alongside
  backend services dedicated to processing, scoring, and providing
  adaptive feedback. This setup enhances gamification elements such as
  progress tracking, achievement systems, and adaptive feedback,
  supporting sustained high motivation levels during learning. The React
  components handle all user interactions, including the welcoming
  chatbot dialogue, writing submissions, adaptive feedback cards, and
  scoring. Meanwhile, the backend services offer endpoints that
  integrate NLP models responsible for scoring and adaptive feedback
  algorithms. The separation of frontend and backend, along with the web
  application\textquotesingle s design, maximises system scalability,
  fault tolerance, and future maintainability, creating a comprehensive
  IELTS preparation experience for candidates.
\end{enumerate}

\hypertarget{b.-conversational-dialogue-component-analysis}{%
\subsection{B. Conversational Dialogue Component Analysis}\label{b.-conversational-dialogue-component-analysis}}

The conversational dialogue guides candidates into the writing
interface, deliberately creating a highly task-focused environment by
breaking down the writing task with a preview card that displays the
initial sections of the IELTS writing task (introduction, body, and
conclusion). This increases user engagement and familiarity with the
platform. Conversational dialogues have already proven successful within
linguistic platforms such as ELSA Speak, establishing
low-friction, learner-friendly environments that foster high engagement
and motivation {\cite{khoi2024investigating}}.

\begin{figure}[H]
\centering
\includegraphics[width=0.9\linewidth]{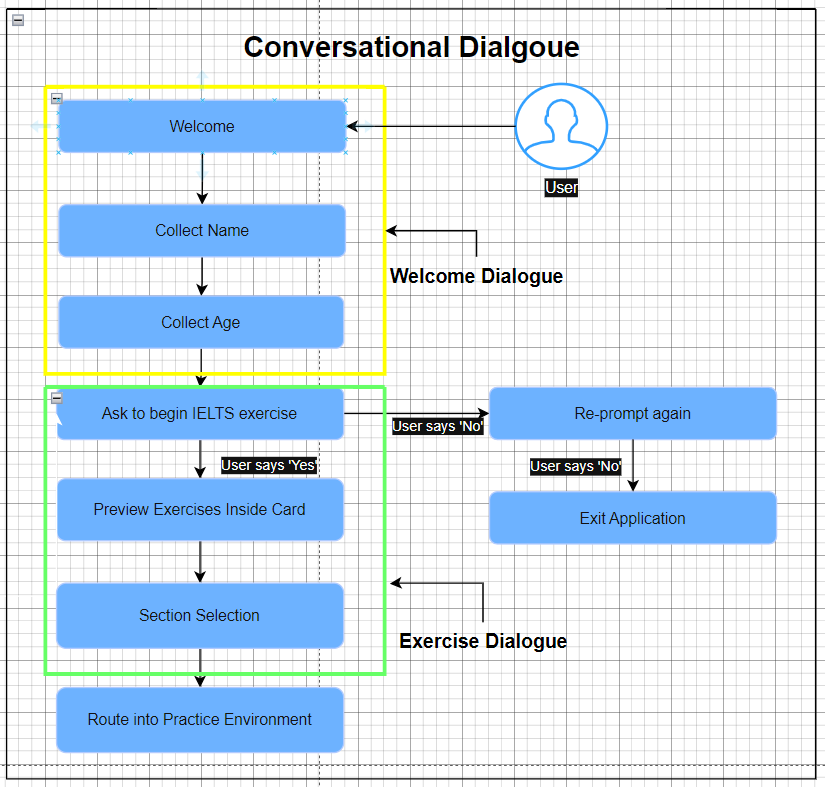}
\caption{Conversational Dialogue Architecture.}
\label{fig:platform}
\end{figure}

Fig. 2 illustrates the main dialogue architecture of the chatbot
designed to guide users into the writing exercise ecosystem, created
using the MBF Azure chatbot. The dialogue begins when a user loads the
system, followed by a greeting and prompts for their name and age,
simulating a typical human conversation. Framing interactions in a
familiar manner, initiating interaction with the linguistic platform.

After completing the welcome dialogue, the exercise flow begins. The
chatbot asks if the user wants to start an IELTS exercise. If the user
agrees, a preview card shows the exercise options, allowing them to
select a writing section. The selected section then guides the user into the main practice environment. This
flow keeps users engaged by breaking the interaction into focused,
manageable parts, helping them stay alert and involved throughout the
session.

MBF's architecture provides leeway within user responses, aiding in constructing an education chatbot
specifically for IELTS preparation, delivering interactive 24/7 learning
support, aspects crucial for a valuable chatbot {\cite{huang2022chatbots}}.

\hypertarget{c.-adaptive-feedback-mechanism-analysis}{%
\subsection{C. Adaptive Feedback Mechanism Analysis}\label{c.-adaptive-feedback-mechanism-analysis}}

The adaptive feedback mechanism distributes feedback catered to candidate writing strengths and weaknesses compared to the IELTS writing
assessment rubric: TA, CC, LR, and GRA. Through rubric decomposition, the
feedback mechanism is better able to pinpoint candidate strengths and
weaknesses. Enabling a
cyclical improvement loop, steering away from non-directional self-assessment.

Adaptive feedback provides two main advantages. First, it helps maintain
high motivation and engagement by offering iterative improvements based on user performance. Secondly, it enhances traditional
feedback by delivering customised guidance that can help improve writing
scores. Embedded within the writing exercises, adaptive feedback
promotes ongoing improvement and keeps users focused throughout
practice sessions.

\begin{figure}[H]
\centering
\includegraphics[width=0.9\linewidth,height=12cm]{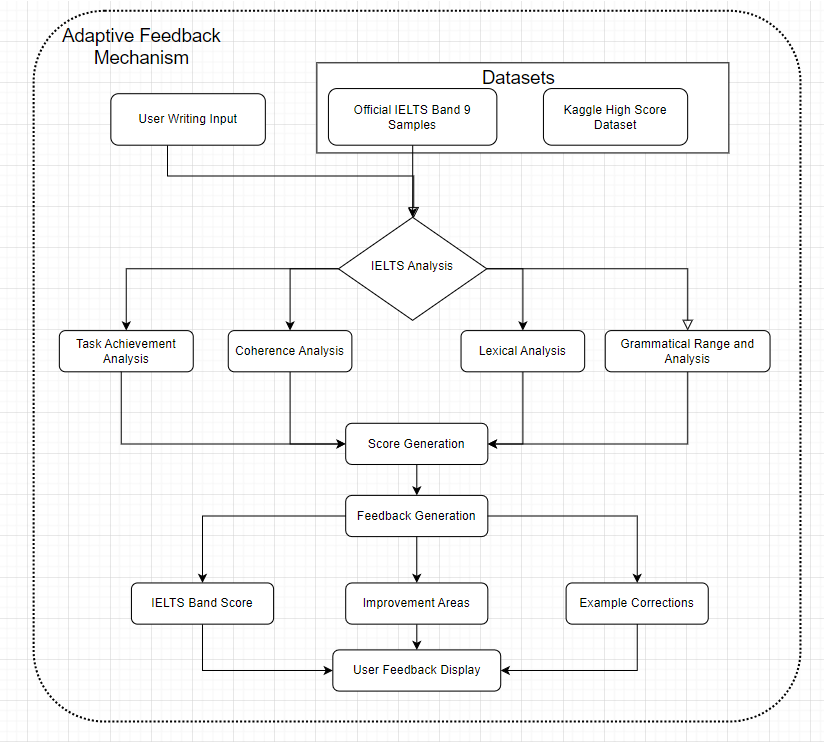}
\caption{Adaptive Feedback Mechanism Architecture.}
\label{fig:adaptive_feedback}
\end{figure}

Fig. 3 illustrates how user submissions are processed via the adaptive
feedback system, which provides predicted band scores, areas for
improvement, and sample corrections. This mechanism is built on official
IELTS samples and human-rated Kaggle datasets, which train a NLP model to deliver precise predictions and analysis in alignment with the scoring rubric.

Research by Reyes is supportive of including an adaptive
feedback system, demonstrating that
algorithm-driven adaptive teaching enhances motivation, engagement, and
performance {\cite{reyes2024exploring}}. In the classroom, adaptive feedback addresses the
lack of personalised learning paths common in traditional methods,
offering consistent, accessible, and formative support instead {\cite{schipper2023dealing}}.
Linguistic platforms with adaptive feedback have historically
experienced higher retention rates.

Adaptive feedback is imperial for effective IELTS preparation. It
identifies gaps and readjustments based on user
performance, guiding candidates to improve their scores according to the
IELTS rubric. This approach helps candidates understand key writing
qualities such as structure, vocabulary, and grammar in an exam context
{\cite{yin2023chatbot}} {\cite{sajja2024artificial}}.

Currently, no existing linguistic platform exactly replicates the IELTS
examiner rubric, which highlights the importance of an IELTS writing practice platform.

\hypertarget{d.-writing-interface-component-analysis}{%
\subsection{D. Writing Interface Component Analysis}\label{d.-writing-interface-component-analysis}}

The IELTS writing interface offers candidates an undisturbed self-study platform reducing cognitive load. 

\begin{figure}[H]
\centering
\includegraphics[width=1\linewidth]{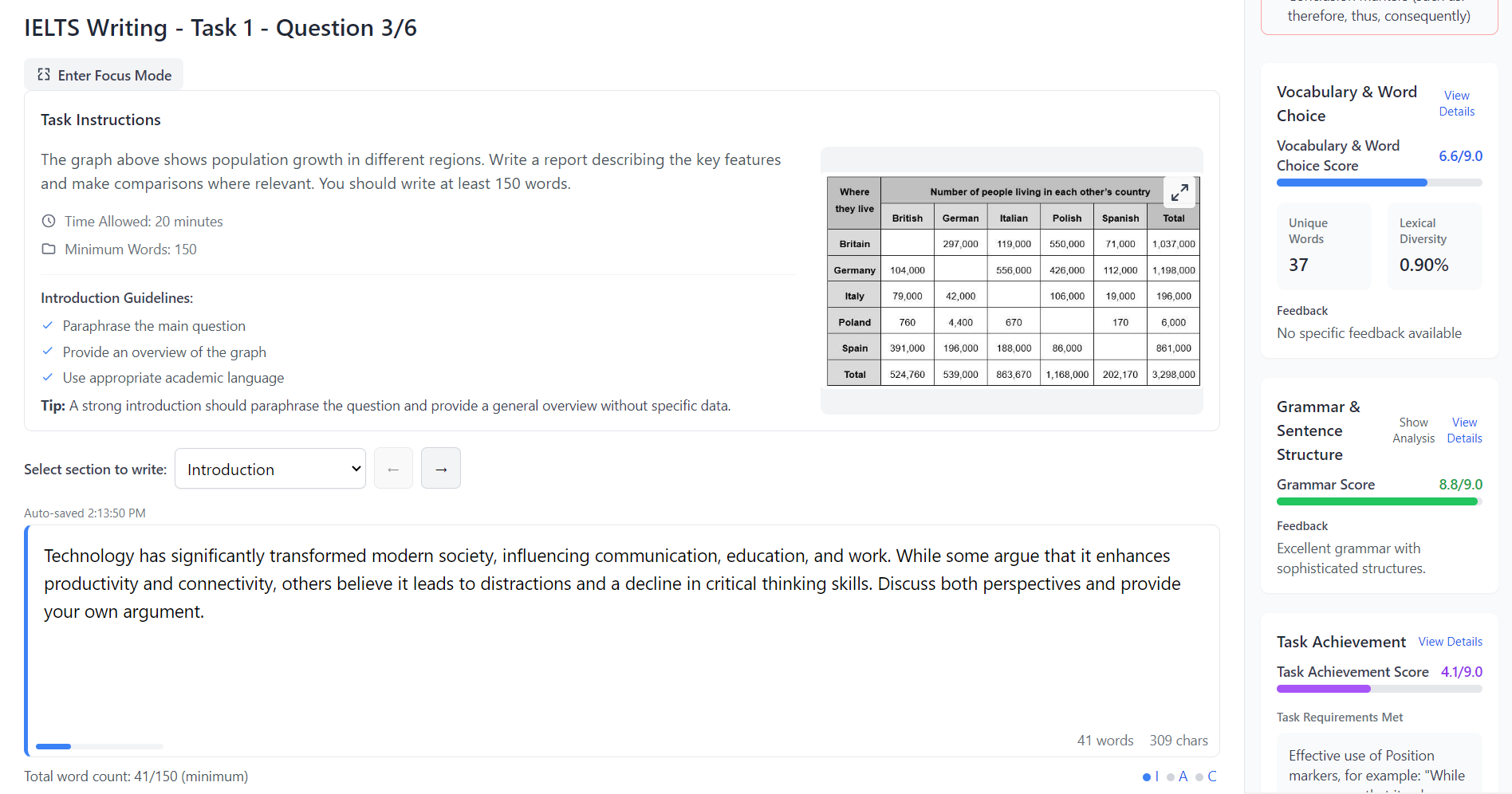}
\caption{IELTS Revision Platform Writing Interface.}
\label{fig:platform}
\end{figure}

As shown in Fig. 4, the interface features task instructions accompanied
by a reference image, a spacious writing area, and a dismissible
sidebar. The sidebar manages submissions, tracks progress, and provides
adaptive feedback, creating a loop for skill improvement. By breaking
down tasks and providing a structured layout, the interface decreases
anxiety.

The success of the writing interface relies on cognitive load theory and
visual hierarchy principles, which improve usability. Fang and
Bueno-Vesga\textquotesingle s research shows that unorganised interfaces
increase mental strain, particularly targeting working memory limits,
leading to lower engagement and retention rates, especially for
second-language learners {\cite{fang2025effects}} {\cite{buenovesga2021effects}}.

Mohamed\textquotesingle s study supports that a clear layout with
deliberate spacing and navigation enhances user attention and
understanding, and wireframing beforehand ensures system functionality
with minimal friction {\cite{mohamed2025effect}}. Limpanopparat\textquotesingle s research indicates engagement peaks with
intuitive navigation when employing tactics such as objective-aligned
content and responsive feedback mechanisms {\cite{limpanopparat2024user}}.

These factors led to choosing a dedicated writing interface over a
chatbot for IELTS prep, as it offers higher immersion, motivation, and
persistence.

\hypertarget{iv.-system-implementation}{%
\section{IV. System Implementation}\label{iv.-system-implementation}}

Our motivation for creating an IELTS preparation platform arises from
the need to offer English learners realistic and progressive practice
opportunities. Traditionally, IELTS preparation has focused on
conventional classroom settings or self-study using online resources.
However, this approach has resulted in a lack of targeted, actionable
feedback and time-limited support, which can lead to poor exam
performance.

Ongoing issues such as difficulty receiving criteria-related feedback,
ineffective self-study, and heightened anxiety during exams prompted the
development of this platform. The platform was designed with four clear objectives outlined in Table IV.

\begin{table}[H]
\centering
\caption{Proposed System Goals and Core Functionalities}
\begin{tabular}{p{5cm} p{10.5cm}}
\toprule
\textbf{Goals} & \textbf{Core Functionalities} \\
\midrule
\addlinespace[0.5em]
IELTS Exam Simulation & 
Realistic conditions (timed task, task format, dedicated input field) \\
\addlinespace[0.5em]

IELTS Writing Rubric Aligned Adaptive Feedback & 
Automated, accurate scoring paired with actionable feedback into Task Achievement, Coherence and Cohesion, Lexical Resource, and Grammar \\
\addlinespace[0.5em]

Minimal Cognitive Load through UI/UX best practices & 
UI/UX incorporating visual hierarchy principles, channelling candidate focus to the writing task \\
\addlinespace[0.5em]

Guarantee Easy Expansion through an isolated, modular system architecture & 
Separation of concerns (React frontend, Chatbot middleware, Python backend), enabling distributed deployment \\
\addlinespace[0.5em]
\bottomrule
\end{tabular}
\label{tab:system_goals}
\end{table}

\begin{enumerate}
\def\labelenumi{\arabic{enumi}.}
\item
  \emph{System Performance:} To maintain high and stable student
  engagement, it is essential to optimise performance. An
  unresponsive system can compromise learners' focus, motivation, and
  may discourage further study. This is especially true for our adaptive
  feedback mechanism, which processing candidate submissions using
  intensive computations. Therefore, to ensure a highly efficient system, performance
  optimisation techniques such as asynchronous operations and threading
  must be employed. Research by Qun on retention rates in Chinese EFL
  learners further supports this prioritisation of system performance,
  linking immediate feedback to high student engagement and performance {\cite{qun2025investigating}}. Conversely, delayed feedback encourages deeper, longer reflection, which is less suited to
  our system.
\item
  \emph{Extensibility:} The system must be adept for expansion; therefore, adherence to best practices is vital across the platform's codebase. Within our platform, best practices are
  present via separation of concerns, where the React frontend, the
  adaptive feedback mechanism, and the chatbot are all loosely coupled
  components. The platform benefits from the best practices being
  followed, enabling the addition and or removal of features independent
  of the entire system {\cite{nivedhaa2024software}}. Granting our platform the ability to
  evolve and adapt over time in response to the feedback provided by candidates.
\item
  \emph{Scalability:} Alongside Extensibility, our system must maintain
  performance as user numbers grow, while ensuring core functionalities
  like UI rendering, adaptive feedback, and automated scoring remain
  fast. Addressing scalability challenges is essential to maximise the
  value delivered to candidates {\cite{alfarra2024ai}}.
\end{enumerate}

\begin{figure}[H]
\centering
\includegraphics[width=0.8\linewidth,height=10cm]{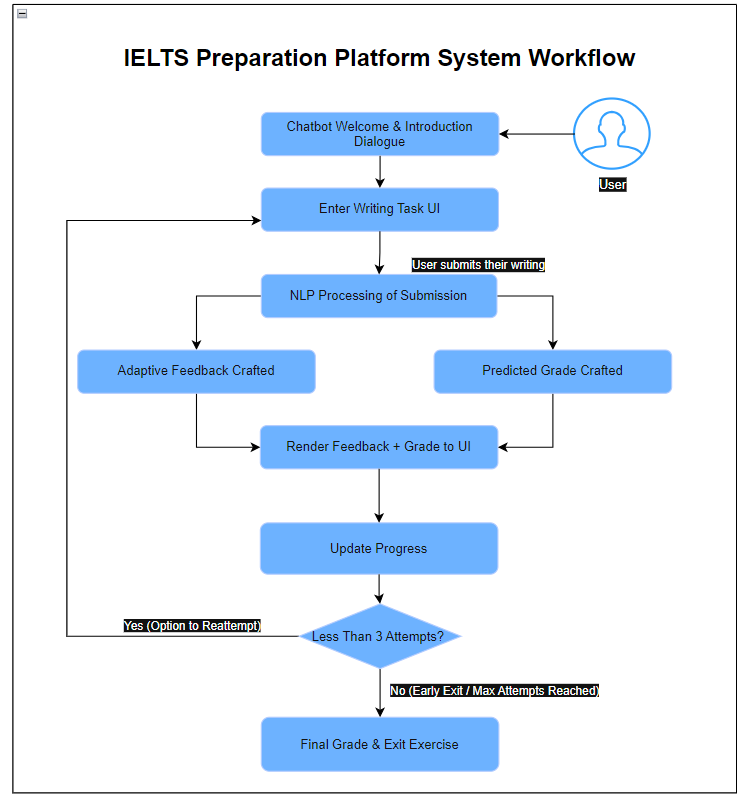}
\caption{Revision Platform Workflow.}
\label{fig:platform}
\end{figure}

Fig. 5 shows the system workflow diagram, which presents core objectives
from Table IV in a logical, user-friendly way, prioritising the
user\textquotesingle s experience. The user is welcomed, followed by an interactive writing task interface 
simulating an examination, featuring tasks, hints, and a dedicated
writing area, meeting a core objectives outlined in Table IV.

After the user successfully submits their attempt, it is assessed
against the IELTS rubric criteria, generating a predicted band score
(1-9) and a grading percentage (0-100\%), along with feedback specific
to the submission, including strong points, weak points, and
suggestions. 

Our linguistic platform is designed for distributed deployment, with a
focus on maintaining high responsiveness and scalability. We use Docker
to isolate and contain NLP components, making them accessible and
interactive throughout the UI utilising asynchronous messaging
queues. This approach ensures low latency for users and quick access to
adaptive feedback, automated grading, and submission times. By doing so,
we can reduce excessive NLP workflows, maintain low latency, and
accommodate scaling as our user base grows.

The UI presents NLP results via modal and sidebar components, including
overall grade, attempts left, exit option, and adaptive feedback
preview, all organised to reduce cognitive load and focus the user on
the writing task. It enforces a three-attempt limit to improve
submission quality and reduce NLP processing load, automatically
progressing the user or ending with final grading and feedback.

\begin{table}[H]
\centering
\caption{System Implementation Specification}
\begin{tabular}{p{2.85cm} p{3.5cm} p{2.5cm} p{6cm}}
\toprule
\textbf{Component} & \textbf{Technology} & \textbf{Version} & \textbf{Purpose} \\
\midrule
\addlinespace[0.5em]

User Interface & React & v18.2+ & 
Rendering UI components and managing user interactions. \\
\addlinespace[0.5em]

Conversational AI & Microsoft Bot Framework & SDK v4.22.0 \& .NET 6 & 
Chatbot dialogue for onboarding and navigating towards the writing exercise. \\
\addlinespace[0.5em]

System Backend & FastAPI \& Python & v3.10 & 
REST API endpoints, Adaptive Feedback Retrieval, and database integration. \\
\addlinespace[0.5em]

NLP Processing & BERT (Transformers) & v4.x & 
Text comparison against the IELTS mark scheme and adaptive feedback generation. \\
\addlinespace[0.5em]

Relational Database & PostgreSQL & v14+ & 
Store user progress, submissions, results, and feedback. \\
\addlinespace[0.5em]

Analytical Database & Firebase & v13.1+ & 
Collect analytics and interaction logs for feature evaluation. \\
\addlinespace[0.5em]

Cloud Hosting & Azure/AWS & Latest & 
Scalable hosting with load balancing and autoscaling. \\
\addlinespace[0.5em]

\bottomrule
\end{tabular}
\label{tab:system_specification}
\end{table}

Table V details the technologies and versions for system
reproducibility and to encourage future project development. Strategies
currently cemented into the platform include distributed and
heterogeneous workloads that are managed to ensure the capability of
handling demanding workloads, such as asynchronous operations to
optimise NLP pipelines, and common practices within high-performance
computing (HPC) environments {\cite{dakic2024evolving}}. Additionally, decoupling
components through leveraging a modular architecture throughout the
codebase improves maintainability, extensibility, and scalability
{\cite{mbugua2022software}}.

Our system will undergo assessments of technical and educational
metrics. Technical evaluations entail stress testing and performance
measurements, spotlighting areas such as latency, scalability, and
system reliability to determine whether the
system can simulate IELTS practice under realistic workloads.
Educational evaluations assess the credibility of the IELTS learning and
preparation benefits provided to candidates. This will be done by
benchmarking the automated grading, actionable insights, and adaptive
feedback. Additionally, the interface will be aligned with UX/UI
principles for pedagogical purposes, primarily reducing cognitive load
for users.

Although user studies are currently outside the scope, our evaluation
strategies will still highlight the platform\textquotesingle s
reliability and educational validity. The methodology and experimental
procedures will be discussed in more detail in Section V (Methodology).

\hypertarget{v.-methodology}{%
\section{V. METHODOLOGY}\label{v.-methodology}}

This study conducts a design-based research (DBR) approach to
develop, test, and refine a linguistic platform specific to IELTS writing preparation. As communicated by Collins, DBR unifies design and research, enabling crafting technical artefacts while investigating their educational impact {\cite{dede2005design}}.

Technical evaluation adhered to performance and load testing principles for web applications{\cite{deep2024performance}},
including unit, end-to-end, latency, and benchmark testing of the
automated grading, user interface, and adaptive feedback features using mimicking candidate submissions.

Through the combination of software engineering with pedagogical evaluation, DBR ensures the platform\textquotesingle s technical reliability and educational validity. Aligning UI designs with best
practices such as visual hierarchy {\cite{mohamed2025effect}}, cognitive load reduction
{\cite{buenovesga2021effects}}, and applying principles of self-determination theory (SDT)
{\cite{laksanasut2025gamification}} to maximise candidate problem-solving capabilities.

\hypertarget{a.-dbr-cycle-1}{%
\subsection{A. DBR Cycle 1}\label{a.-dbr-cycle-1}}

The initial DBR cycle aimed to establish the foundational modular architecture of the IELTS preparation platform. This involved developing early prototypes of the UI, MBF chatbot, and Python NLP-based automated
grading and adaptive feedback key platform components.   

The visual hierarchy of the UI was designed based on principles blending both cognitive load theory {\cite{fang2025effects}} and SDT {\cite{laksanasut2025gamification}} to optimise
candidate focus, clarity, reduce mental load, and reward candidate
effort. This approach ensured the platform was clear, accessible, and user-friendly, thereby encouraging greater motivation to study.

The first DBR cycle resulted in a working prototype capable of
collecting user input, grading responses, and providing adaptive
feedback. This laid a strong technical and pedagogical foundation for
subsequent DBR cycles.

The second DBR cycle assessed the technical accuracy and pedagogical
consistency of the IELTS preparation platform. Since the platform
includes an automated feedback system, it was crucial to ensure that its
outputs align with IELTS examiners\textquotesingle{} standards to
provide candidates with trustworthy guidance {\cite{shermis2013handbook}}.

\hypertarget{b.-dbr-cycle-2}{%
\subsection{B. DBR Cycle 2}\label{B.-dbr-cycle-2}}

The diagnosis strategy covered three testing areas. First, unit testing
using Jest {\cite{jest2025}}, Pytest {\cite{pytest2025}}, and xUnit {\cite{xunit2025}} checked
components like chatbot dialogues, user interface, and grading services.

Secondly, end-to-end testing simulated user workflows using Cypress
{\cite{cypress2025}} from chatbot greeting to submission, ensuring smoothness.
Lastly, benchmarking assessed NLP model predictions against a Kaggle
IETS dataset {\cite{kaggle2025ielts}}, measuring accuracy and discrepancies.

This cycle exposed limitations in strict grammar grading,
inconsistencies in LR, and underestimating higher-band
essays.

These findings helped guide the subsequent cycle, focusing on remediating
grading and feedback methods, aligning with IELTS exam standards, and
enhancing pedagogical effectiveness.

\hypertarget{c.-dbr-cycle-3}{%
\subsection{C. DBR Cycle 3}\label{C.-dbr-cycle-3}}

The third DBR cycle focused on implementing fixes for weaker areas identified in the cycle 2. Remediation targeted three regions. First, the CoLA grammar model {\cite{potapczynski2023exploiting}}, which was less
consistent in analysing and identifying errors, was replaced with
Python's LanguageTool API {\cite{languagetool2025}}, providing evaluations that adhere more closely to the IELTS Grammatical Accuracy and Range rubric.

This provided more IELTS-appropriate evaluations within the GRA criteria {\cite{languagetool2025}}. Second, the LR classification process was remodeled to incorporate word sophistication
ratios {\cite{kyle2018tool}} instead of relying solely on academic vocabulary
frequency.

Lastly, the task achievement grading algorithm was reformulated to
mitigate the issue of low scores assigned when submissions did not meet
the 250-word count requirement, introducing a linear scaling model to
grade submissions up to the word limit.

To gauge remediation effectiveness, the automated grading system was re-evaluated using benchmark testing on 100 pre-scored essays from the Kaggle IELTS dataset {\cite{kaggle2025ielts}}.

Performance metrics included mean absolute error (MAE), coefficient of
determination ($R^2$), and the percentage of predictions within ±0.5 IELTS
bands of expert scores.

\hypertarget{d.-dbr-cycle-4}{%
\subsection{D. DBR Cycle 4}\label{d.-dbr-cycle-4}}

The fourth DBR cycle addressed the limitations of rule-based and hybrid
scoring methods for grading IELTS writing essays.

While cycle 3 improved scoring for GRA, LR, and TA, grading was still not reflective of the full range. There were underpredictions of higher-band essays and a tendency to cluster grades around the middle bands. Investigation revealed that the rule-based AES system could be deceived by longer essays and complex vocabulary, exposing its poor generalisation, an issue common to many AES algorithms {\cite{ifenthaler2022automated}}.

In response, Cycle 4 shift to a lightweight transformer-based model with a regression head, specifically DistilBERT, to enhance predictive accuracy, guided by previous research led by Ludwig investigating transformers\textquotesingle{} suitability for AES {\cite{ludwig2021automated}}.

The choice to adopt a Transformer architecture stemmed from the underperformance of previous rule-based approach and the researched advantages of pre-trained language models in AES. Traditional models relying on features like word counts and sentence length often struggle to capture complex writing styles, such as rhetorical structures or medium-level techniques {\cite{ludwig2021automated}}.

Studies show that transformer models outperform regression-based methods, especially for tasks involving style, coherence, and context-areas difficult for simpler approaches {\cite{ludwig2021automated}}. This improvement directly influenced DBR Cycle 5, which focused on implementing adaptive feedback. Accurate predictions are crucial; inaccurate scores could lead to misleading feedback, hindering learners\textquotesingle{} progress and damaging the platform\textquotesingle s credibility.

Each essay is tokenised using the standard DistilBERT tokeniser and is
either truncated or padded to a fixed maximum length of 256 tokens. This
process ensures consistent input size while preserving most of the
essay\textquotesingle s content.

The resulting tokens are fed into a pretrained encoder to generate contextualised embeddings; the pooled representation is then concatenated with a vector of manually extracted linguistic features (such as word count, sentence length, lexical diversity, and punctuation density), which are modelled through our iterations in DBR Cycles 2 and 3. Research indicates that hybrid models combining deep embeddings with
explicit linguistic indicators often outperform models that rely solely on features or embeddings {\cite{faseeh2024hybrid}}.

A linear regression head on top of this combined representation produces
a continuous score predicting the IELTS band. Framing band prediction as
a regression, rather than a classification, maintains the ordinal and
constant nature of the scoring system and has been shown to improve
calibration in automated scoring systems {\cite{misgna2024survey}}.

To balance expressive power and generalisation, the model is trained
using the AdamW optimiser with a learning rate of $1.5 \times 10^{-5}$, weight
decay of 0.02, and batch-size adjustments via gradient accumulation to
simulate an effective larger batch while remaining memory efficient.

Dropout rate of 0.35, gentle label smoothing, and partial freezing of
the encoder layers are applied to mitigate overfitting, a common
challenge when fine-tuning large pretrained models on modestly sized
essay datasets. Early stopping on validation MAE with patience
monitoring ensures the selection of a model that generalises beyond
training data. This balanced regularisation regime follows best
practices identified in AES research using Transformer models, ensuring
robust performance without excessive memorisation {\cite{mayfield2020finetune}}.

The dataset was randomly partitioned into 70\% training, 15\%
validation, and 15\% test, provisioning reliable hyperparameter tuning
while preserving an independent final evaluation {\cite{kaggle2025ielts}}. The training
was run for up to 30 epochs, with early stopping based on validation MAE
to prevent overfitting {\cite{wikipedia2025earlystopping}}

To evaluate cycle 4's scoring model, we followed AES research practices.

We assessed Mean Absolute Error (MAE) for average deviation, $R^2$ for variance explained, and accuracy within ±0.5 and ±1.0 IELTS bands {\cite{xu2024systematic}}.

We also used Pearson's correlation coefficient (PCC) and Spearman's ($\rho$)
to measure linear and monotonic relationships {\cite{xu2024systematic}}. $\rho$ is especially
important for ordinal IELTS bands, as it checks whether the model preserves the scoring order.

These metrics collectively provide a clear overview of accuracy,
distribution, and order preservation, enabling transparent assessment of
the Cycle 4 model.

\hypertarget{e.-dbr-cycle-5}{%
\subsection{E. DBR Cycle 5}\label{e.-dbr-cycle-5}}

The fifth DBR cycle centred on implementing and validating adaptive
feedback based on the transformer- based grading model from Cycle 4.

While earlier cycles focused on achieving grading accuracy and aligning
with rubrics, Cycle 5 explored if algorithmic feedback could facilitate
meaningful score improvements through simulated candidate revisions
{\cite{fleckenstein2023automated}}.

To prevent data leakage and ensure fair assessment, a subset of 30 IELTS
Writing Task 2 essays was selected from an external dataset not used in
training, validation, or testing during Cycle 4. These essays were
sourced from a publicly available IELTS evaluation corpus {\cite{chi2023huggingface}},
ensuring they were unseen data.

The transformer model from Cycle 4 was retained as a fixed scoring
component during Cycle 5, so score changes could be attributed solely to
the feedback mechanism, not model retraining or drift. Adaptive feedback
generation reused algorithms from Cycles 2 and 3, with prioritisation
and targeting now guided by the predicted band score and associated
linguistic feature deviations of the transformer model {\cite{misgna2024survey}}.

An evaluation process consisting of three steps was used to ensure
fairness, reproducibility, and independence from prior training effects.
First, each essay was scored with the transformer- based regression
model, producing a continuous IELTS band prediction used as a baseline.

Second, adaptive feedback was retrieved based on strengths and
weaknesses in IELTS Writing criteria: Grammatical Range and Accuracy,
Lexical Resource, Coherence and Cohesion, and Task Achievement. This
feedback was informed by the predicted score and linguistic deviations,
targeting specific writing aspects likely to improve the band score.

Finally, feedback was applied through candidate persona revision
behaviours, and the revised essays were rescored with the same model and
preprocessing pipeline, mirroring the initial evaluation process. This
closed- loop approach operationalises DBR principles by embedding
theory- informed feedback and immediately assessing its instructional
impact under controlled conditions.

\begin{table}[H]
\centering
\caption{Candidate Revision Personas}
\begin{tabularx}{\textwidth}{p{1.3cm} p{1.3cm} p{2.2cm} p{1.8cm} X}
\toprule
\textbf{Persona} & \textbf{Band Level} & \textbf{Revision Style} & \textbf{Compliance Level} & \textbf{Revision Actions} \\
\midrule
\addlinespace[0.5em]

P1 & 6.5 & Surface edits & High & 
Grammar, spelling, and punctuation fixes. Never restructure essays. \\
\addlinespace[0.5em]

P2 & 6.5–6.8 & Structural & Medium & 
Frequent paragraph splitting. Implement topic sentences. Minimalistic vocab editing. \\
\addlinespace[0.5em]

P3 & 6.8 & Selective & Low–Medium & 
Implements only 1–2 feedback suggestions, ignores the rest of the feedback. \\
\addlinespace[0.5em]

P4 & 7.0 & Coherence Focus & High & 
Improves sentence flow. Applies paragraph logical connectors. Avoidant of fancy vocab. \\
\addlinespace[0.5em]

P5 & 6.7 & Conservative & Medium & 
Safe edits, avoid sentence rewrites or new ideas. \\
\addlinespace[0.5em]

\bottomrule
\end{tabularx}
\label{tab:revision_personas}
\end{table}

To ensure candidate behaviour was controllable and reproducible, Cycle 5
simulated typical IELTS revision strategies using five personas
{\cite{nolte2022creating}}, each with an initial band range, revision style, and
compliance level with the feedback (see Table VI). Six essays from the
locked 30 essay dataset were randomly assigned to each persona, and each
person applied adaptive feedback according to their assigned compliance
and style, clearly outlining permissible edits. This setup allowed for a
systematic evaluation of how different revision tactics influenced
feedback effectiveness without uncontrolled or inconsistent edits.

\hypertarget{vi.-results}{%
\section{VI. RESULTS}\label{vi.-results}}

\hypertarget{a.-benchmarking-metrics-across-dbr-cycles}{%
\subsection{A. Benchmarking Metrics Across DBR Cycles}\label{a.-benchmarking-metrics-across-dbr-cycles}}

Table VII summarises benchmarking results for DBR Cycle 2 and Cycle 3,
measuring post-remediation using statistical metrics to evaluate grading
performance, variance, and deviation against the pre-scored dataset.

\begin{table}[H]
\centering
\caption{Metrics across DBR Cycles 2 \& 3}
\begin{tabularx}{\textwidth}{X c c c}
\toprule
\textbf{Metric} & \textbf{DBR Cycle 2} & \textbf{DBR Cycle 3} & \textbf{$\Delta$ Improvement} \\
\midrule
\addlinespace[0.5em]

Exact Match (\%) & 13\% & 17\% & 4\% \\
\addlinespace[0.5em]

Within $\pm 0.5$ band (\%) & 47\% & 60\% & 13\% \\
\addlinespace[0.5em]

$R^2$ & $-0.0242$ & $-0.0052$ & 0.019 \\
\addlinespace[0.5em]

Mean Average Error (MAE) & 1.27 bands & 1.14 bands & 0.013 bands \\
\addlinespace[0.5em]

\bottomrule
\end{tabularx}
\label{tab:dbr_metrics}
\end{table}

As Table~VII shows, DBR Cycle 2 demonstrated poor grading reliability: only 13\% of predictions matched the exact band, with 47\% falling within $\pm 0.5$ bands. The MAE of 14.10\% and negative $R^{2}$ (-0.0242) indicated the model performed worse than simply predicting the mean score, necessitating significant refinement.

Cycle 3 remediations focused on algorithmic refinement of GRA, LR, and TA evaluation. Exact band matches rose to 17\% (+4\%), and $\pm 0.5$ bands accuracy rose to 60\% (+13\%). MAE improved to 12.64\%, and $R^{2}$ moved closer to zero (-0.0052), though remained negative.

Despite improvements, the model retained a mid-band bias (6.0--6.5), suggesting limited generalisation stemming from its lexical diversity weighting. Importantly, the persistently negative $R^{2}$ in both cycles showed the model performed worse than simply predicting the mean score. Though MAE dropped 1.46\%, it remained above acceptable thresholds for accurate assessment, exposing limitations in the rule-based architecture.

We then visualised the predictions versus actual scores (scatter plots, confusion matrices, and box plots) to encapsulate remediation effects. 

\hypertarget{b.-visual-benchmark-analysis-of-dbr-cycle-2}{%
\subsection{B. Visual Benchmark Analysis of DBR Cycle 2}\label{b.-visual-benchmark-analysis-of-dbr-cycle-2}}

\begin{figure}[H]
\centering
\includegraphics[width=1\linewidth,height=7cm]{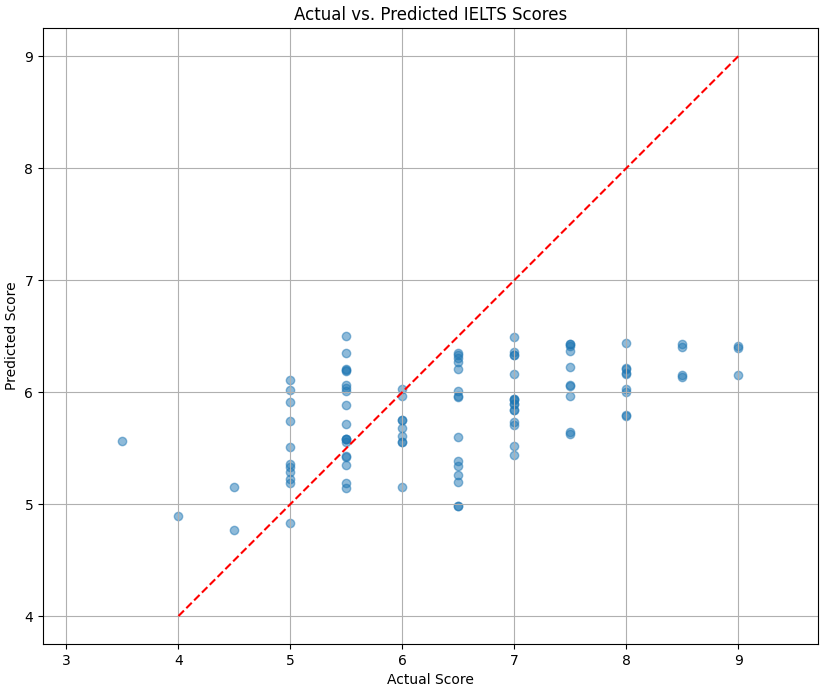}
\caption{(Cycle 2) Scatter plot of actual vs predicted IELTS band
scores.}
\label{fig:platform}
\end{figure}

Most predictions cluster around the 5.5-6.5 range, evidencing that the
model is reliable for mid-range essays.

However, the model tends to underpredict essays around the 7.5-9.0
range, as points fall below the diagonal.

\begin{figure}[H]
\centering
\includegraphics[width=1\linewidth,height=8cm]{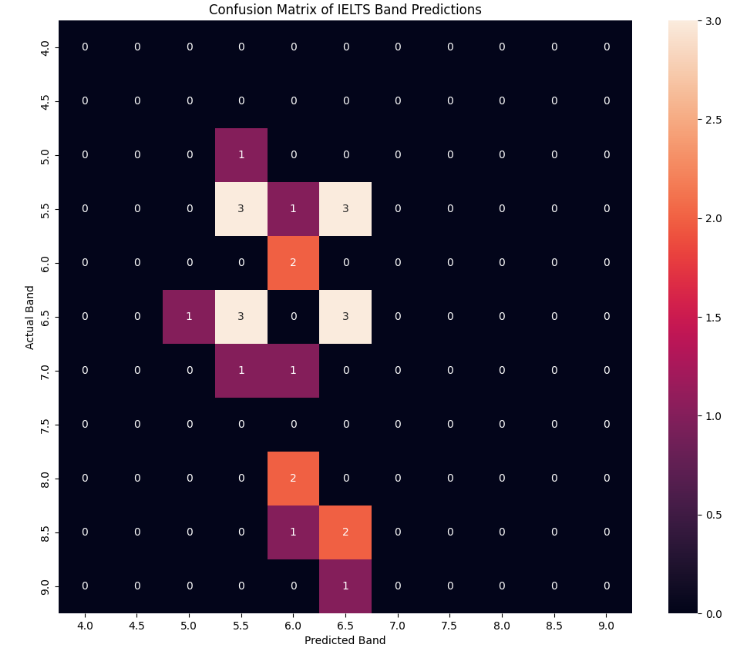}
\caption{(Cycle 2). Confusion matrix of actual vs predicted bands.}
\label{fig:platform}
\end{figure}

Most predictions ranged from 5.5 to 6.5 bands, indicating that the model
does not fully capture the entire score distribution as needed.

\begin{figure}[H]
\centering
\includegraphics[width=1\linewidth,height=7cm]{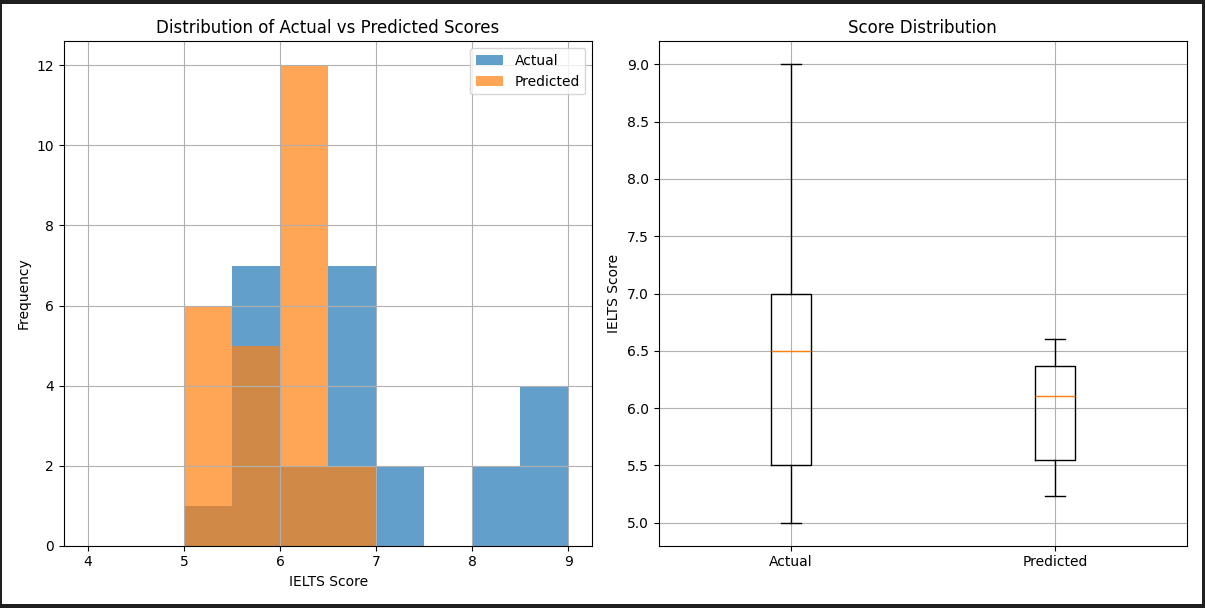}
\caption{(Cycle 2). Box plots of actual vs predicted scores.}
\label{fig:platform}
\end{figure}

The actual scores span from 5.0 to 9.0, whereas the predicted scores are
grouped tightly around 6.0-6.5, underscoring the model's inability to
capture the true dataset variation.

Holistically, Figs 6-8 establish that the Cycle 2 model iteration is
only partially reliable, grading mid-band essays convincingly, but
inconsistently grading higher-band submissions. These findings motivated
the Cycle 3 remediations.

\hypertarget{c.-post-remediation-benchmarking-and-visual-analysis-dbr-cycle-3}{%
\subsection{C. Post-Remediation Benchmarking and Visual Analysis (DBR Cycle
3)}\label{c.-post-remediation-benchmarking-and-visual-analysis-dbr-cycle-3}}

Built on the Cycle 2 analysis, we applied targeted remediations
to GRA, LR, and TA scoring. We substituted the CoLA grammar model with Python's LanguageTool API to obtain a grading better correlated with the IELTS rubric, followed by enriching LR to account for word sophistication.
Finally, we altered the TA grading to incorporate a linear
scaling model, reducing bias against shorter essays. We then re-benchmarked the model on the same 100 pre-scored essays to
fairly evaluate the impact of the remediations performed under identical
conditions.

Table VII confirms advancements were made across all categories. The
exact score matching increased from 13\% to 17\%, and within ±0.5, the
precision increased from 47\% to 60\%. MAE decreased by 1.46\%, and R² became much
less negative.

The superior metrics revealed that the remediations augmented the
model's grading accuracy and reliability.

\begin{figure}[H]
\centering
\includegraphics[width=1\linewidth,height=7cm]{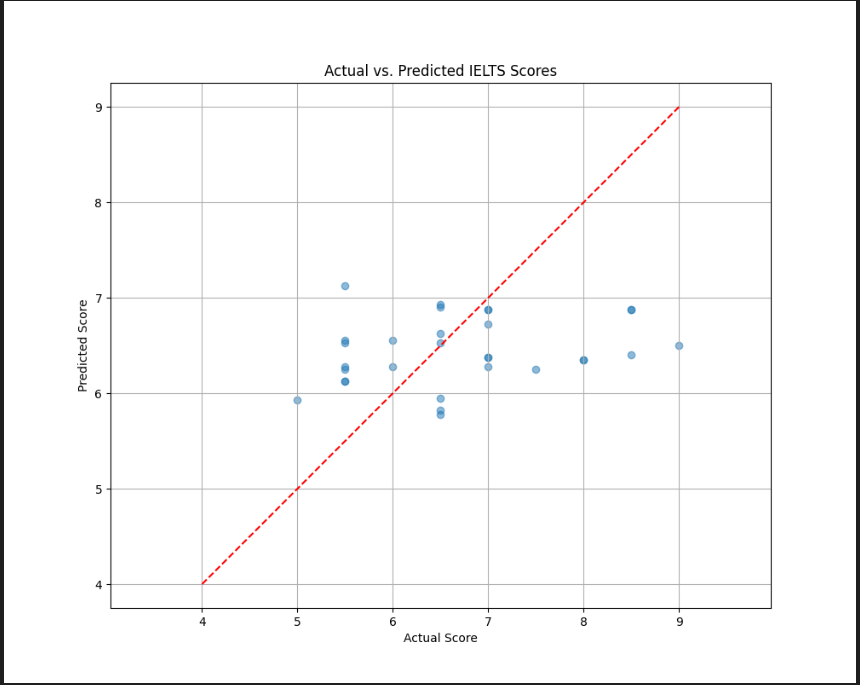}
\caption{(Cycle 3). Scatter plot of actual vs predicted scores post-remediation.}
\label{fig:platform}
\end{figure}

Predictions illustrated scoring closer with actual scores, especially in
the 5.5-7.0 range, validating the recalibrations applied to the GRA,
LR, and TA scoring algorithms. Nevertheless,
the model's underprediction for 7.5-9.0 essays persisted, so further
improvements were required.

\begin{figure}[H]
\centering
\includegraphics[width=0.8\linewidth, height=6cm]{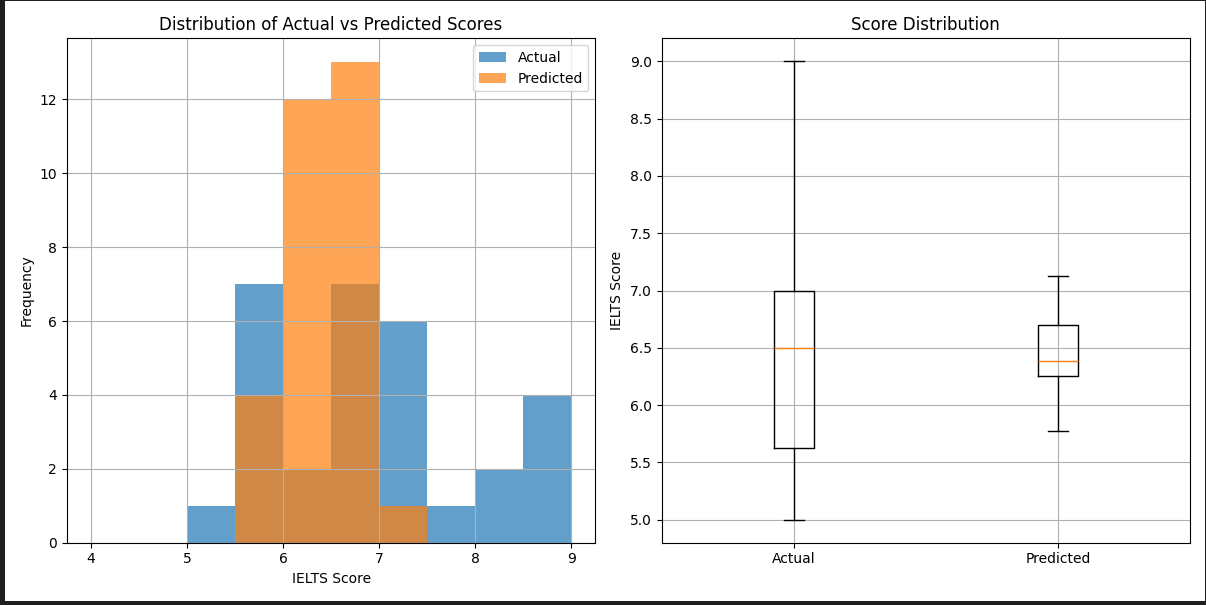}
\caption{(Cycle 3). Box plots of actual vs predicted scores post-remediation.}
\label{fig:platform}
\end{figure}

The predicted range has better coverage of the actual scores after
remediation, yet predictions still group in the 5.5-7.0 range. This
clustering signifies that the model is unable to handle high-quality
(8.0-9.0) essays.

\hypertarget{d.-benchmarking-metrics-for-dbr-cycle-4}{%
\subsection{D. Benchmarking Metrics for DBR Cycle 4 (DistillBERT Regression Model)}\label{d.-benchmarking-metrics-for-dbr-cycle-4}}

Looking back at the results from Cycles 2 and 3, it was apparent that
the rule-based algorithms had a performance ceiling within an AES context, constraining the
accuracy and effectiveness of the adaptive feedback once activated, even
after remediation.

Cycle 4 focused on developing the necessary driver for adaptive feedback
by introducing a lightweight, data-driven regression model based on
DistilBERT. This architecture proved beneficial because it could learn
linguistic and semantic features directly from sample essays, enabling
strong generalisation.

Table VIII presents the performance of these regression models on both
training and testing datasets.

\begin{table}[H]
\centering
\small
\caption{Benchmarking Metrics for DBR Cycle 4}
\begin{tabular}{p{1.6cm} p{1.7cm} p{1.7cm} p{2cm} p{2.5cm} p{2cm} p{2cm}}
\toprule
\textbf{Split} & \textbf{MAE} & \textbf{R\textsuperscript{2}} & \textbf{Pearson (r)} & \textbf{Spearman ($\rho$)} & \textbf{$\pm$ 0.5 (\%)} & \textbf{$\pm$ 1.0 (\%)} \\
\midrule
\addlinespace[0.3em]
Train & 0.605 & 0.495 & 0.711 & 0.682 & 46.9 & 83.7 \\
\addlinespace[0.3em]
Test & 0.666 & 0.354 & 0.601 & 0.577 & 45.2 & 74.8 \\
\addlinespace[0.3em]
\bottomrule
\end{tabular}
\label{tab:dbr_cycle4_metrics}
\end{table}

\hypertarget{e.-metric-analysis-for-dbr-cycle-4}{%
\subsection{E. Metric Analysis for DBR Cycle 4}\label{e.-metric-analysis-for-dbr-cycle-4}}

The changes in Cycle 4 yielded substantial AES improvements compared
with Cycles 2 and 3. The MAE essentially halved to 0.666 on unseen essays (test set), a major
improvement to Cycle 2 and 3's best of 1.14, quantifying an average
predictive accuracy improvement of 0.474 bands.

Crucially, $R^2$ became positive for the first time; the
model correctly explains 35-49.5\% of the data variance in expert-graded
essays, compared with performance below the mean in Cycles 2 and 3,
characterised by negative $R^2$ values.

Additionally, we included the analysis of Pearson metrics to help gauge
the correctness of bands scaling, achieving 0.601-0.711. Indicating that
predictions adhered to the correct numeric spacing, amplifying our MAE
values credibility {\cite{yannakoudakis2018developing}}.

Similarly, Spearman's rank correlation was used to assess the model's
ability to understand ordinal essay characteristics, which was vital for
validly contributing to AES research {\cite{mathias2020neural}}. The Spearman values
(0.577-0.682) suggest that the band-order relationship was captured
reliably, complementing the band-accuracy measures. Both the Spearman and Pearson metrics exceeded 0.60, aligning well with the established benchmarks for AES studies {\cite{xu2024systematic}}.

It should be noted that although the ± 0.5 band accuracy didn't seem to
show much improvement compared to Cycles 2 and 3, the inclusion of the
±1.0 metric aided in painting a true grading performance
indicator. Because Cycle 4 was trained on 1,200 samples and tested on
300 unseen samples, the metrics relayed were more consistent and stable,
and in fact showed far greater generalisability than the 100 sampled
essays in Cycles 2 and 3.

From a benchmarking metric overview, the DistilBERT
regression model improved predictive performance, variance explanation,
and correlation with human scoring. Pillaring a strong podium to begin implementing adaptive feedback into the mix.

\hypertarget{f.-visual-analysis-of-dbr-cycle-4-benchmark-results}{%
\subsection{F. Visual Analysis of DBR Cycle 4 Benchmark Results}\label{f.-visual-analysis-of-dbr-cycle-4-benchmark-results}}

To dissect the improvements observed in Table VI, we generated visual
analytics of the DistilBERT regression model performance.

Figs 11--14 exhibit scatter plots, box-plot distributions, and
confusion matrices across training and testing
datasets, furnishing a deeper understanding about the metrics in Table VIII and the impact of transitioning to a transformer architecture.

\begin{figure}[H]
\centering
\includegraphics[width=0.9\linewidth,height=8cm]{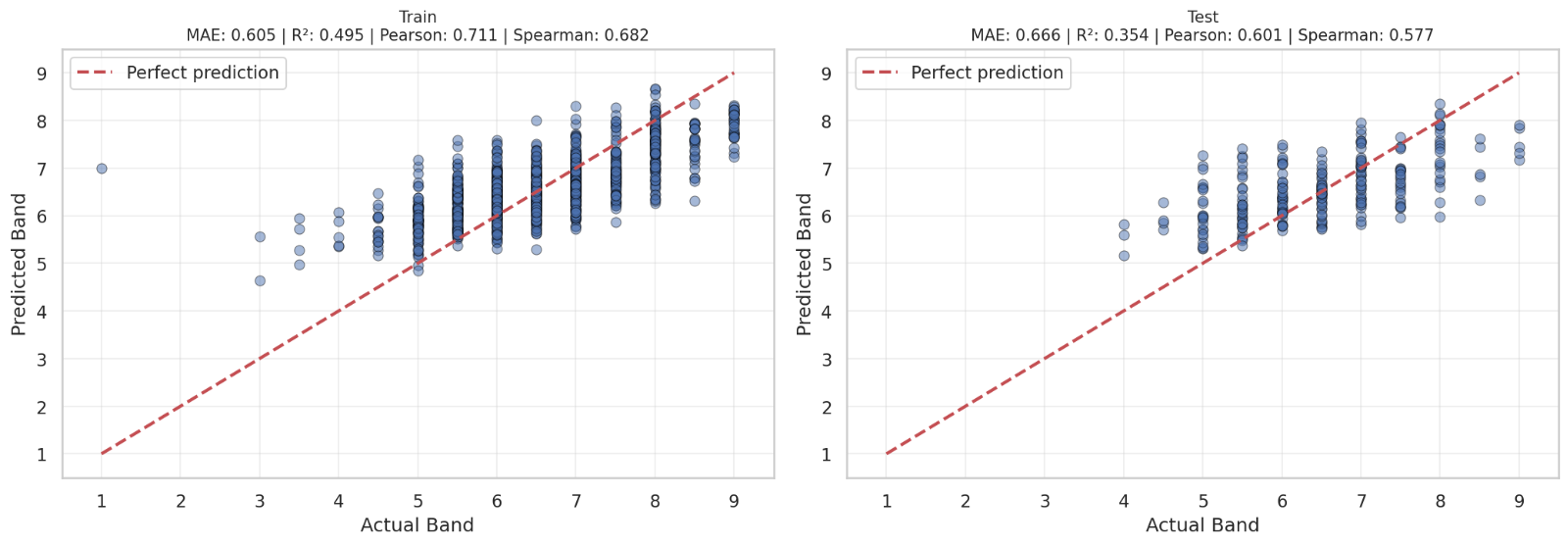}
\caption{(Cycle 4). Scatter plot of actual vs predicted scores (train/test samples).}
\label{fig:platform}
\end{figure}

The predictions were notably more tightly grouped around the line of
best fit compared to Cycles 2 and 3. During training, the predictions
closely followed the best-fit line from bands 4.0 to 7.5, which is
supported by strong Pearson (0.711) and Spearman (0.682) correlations.

Most importantly, the model's scores were more consistent across the
IELTS bands, addressing a pivotal issue with rule-based models that
showed compressed scores in the middle bands.

Predictably, performance slightly declined in the test split, a common
occurrence with unseen data. Nevertheless, the score distribution
remained controlled, with an MAE of 0.666. Higher-band essays (7.5--9.0)
exhibited mild underprediction, but this was significantly less than in
Cycles 2 and 3, where predictions often steeply converged toward bands 6.0--6.5.

Fig. 11 shows a reduced spread, better diagonal alignment, and more
precise scoring coverage, confirming that the transformer architecture
provided scores that more closely resemble expert grading for IELTS
writing.

\begin{figure}[H]
\centering
\includegraphics[width=0.9\linewidth,height=8cm]{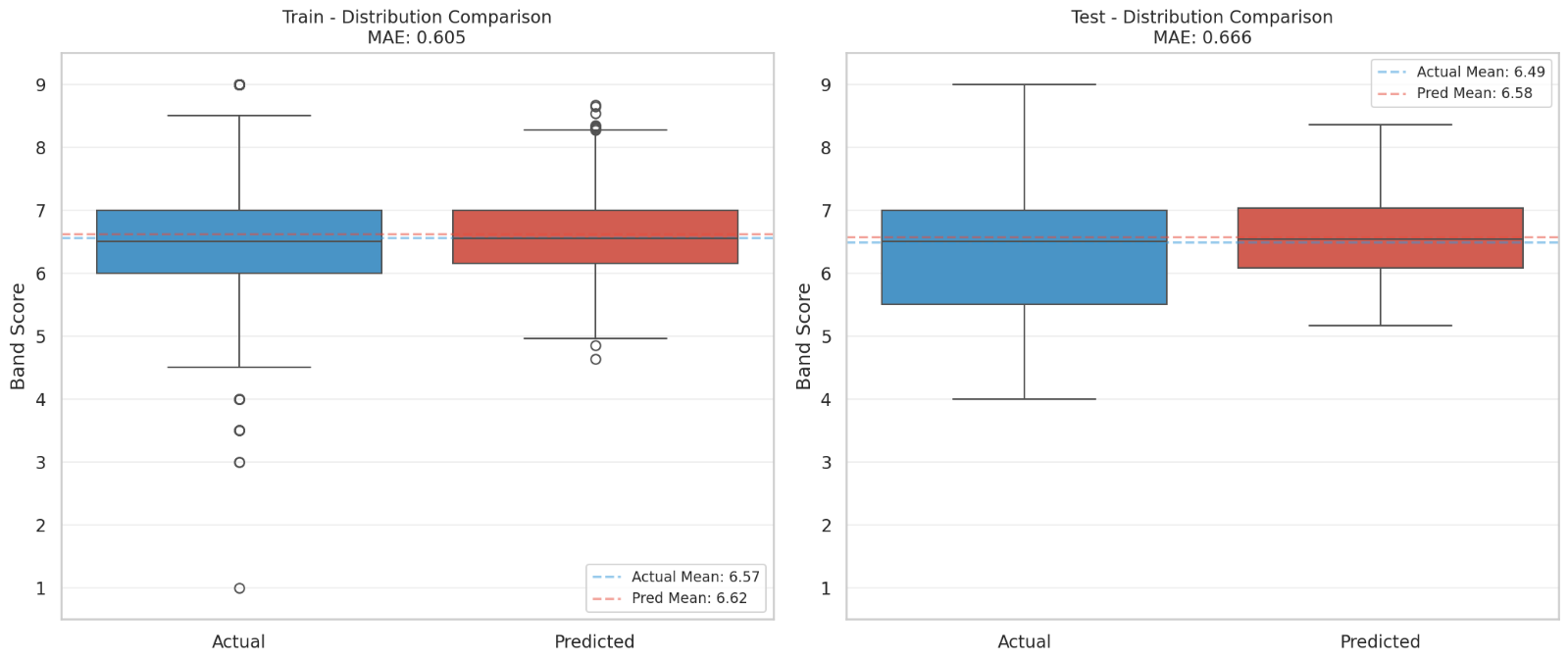}
\caption{Box plots of actual vs predicted scores (train/test samples).}
\label{fig:platform}
\end{figure}

Fig. 12's box plot illustrates the improvement in the
model\textquotesingle s score distribution accuracy following the
upgrade. During training, the model predicted a median score of 6.82,
closely matching the actual median of 6.57, demonstrating the
transformer model's high precision. Its variance is also quite accurate,
with the predicted interquartile ranges mostly overlapping with the
expert-scoring distributions.

For the test set, the model predicted a median of 5.98, slightly
underestimating the actual median of 6.49, but still within an
acceptable range pedagogically.

More importantly, the predicted distribution no longer collapses into
isolated clusters, a key issue that led to replacing the previous model.

Instead, the variance spreads similarly to the real distribution,
confirming that the transformer model effectively generalises linguistic
features and band scores.

The consistency between actual and predicted variance is crucial for
producing intuitive, adaptive feedback. Incorrect clustering would
result in misleading or unstable responses, which could impair IELTS
writing task preparation.

\begin{figure}[H]
\centering
\includegraphics[width=1\linewidth,height=8cm]{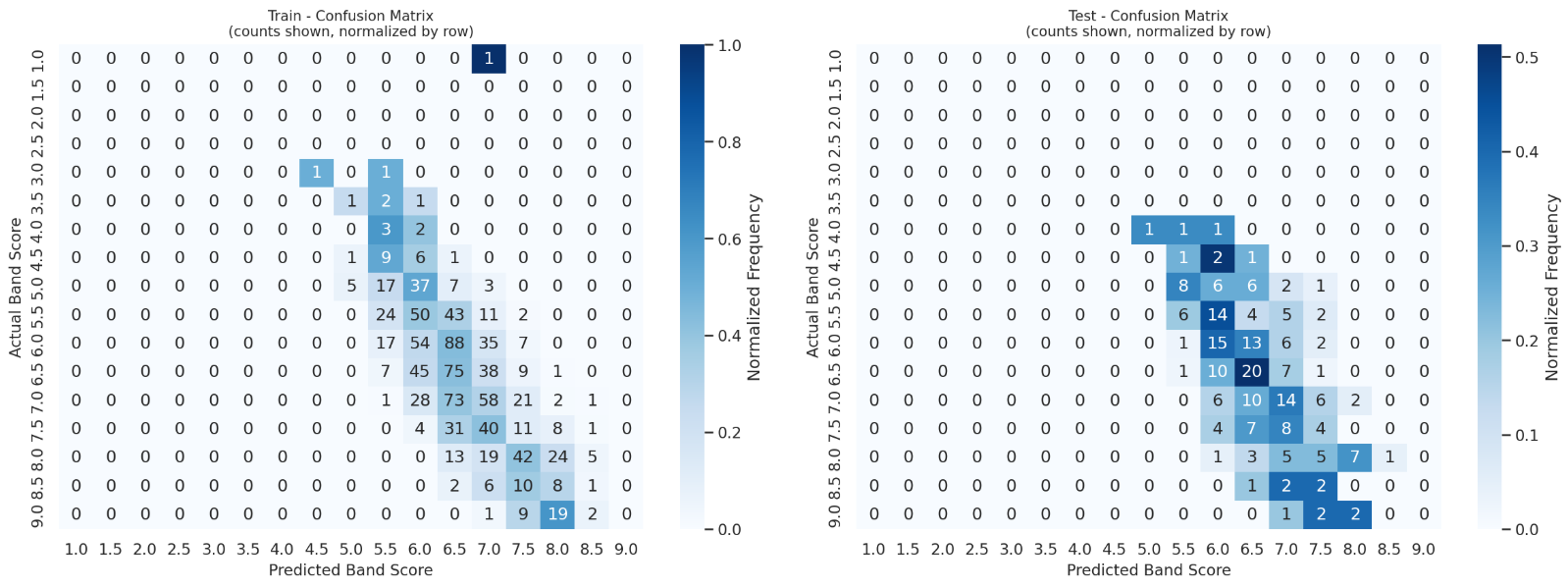}
\caption{(Cycle 4). Confusion matrix of actual vs predicted bands (train/test samples).}
\label{fig:platform}
\end{figure}

The confusion matrices further corroborated the
transformer\textquotesingle s scoring distribution.

Unlike previous cycles, where heat concentration was dense around the
middle bands (5.5-6.5), the transformer\textquotesingle s distribution
remained more consistent with the true band positions, as seen through
the diagonal heat mapping.

Both training and testing matrices indicated most predictions fell
within ±1.0 band of the actual score, reflecting the high 74.8--83.7\%
±1.0 accuracy shown in Table VIII.

Test-set predictions show a slightly wider diagonal band, indicating
modest generalisation variance, but no signs of systematic errors or
bias.

Misclassifications follow expected patterns of IELTS scoring
ambiguities, such as confusing neighbouring bands, rather than errors
from misapplied rules (which occurred in DBR Cycles 2 and 3).

This move toward a more diagonal-dominant matrices is typical of
high-performing AES models. It confirms that the transformer is learning
key hierarchical and semantic features necessary for accurate ordinal
scoring.

\begin{figure}[H]
\centering
\includegraphics[width=0.9\linewidth,height=8cm]{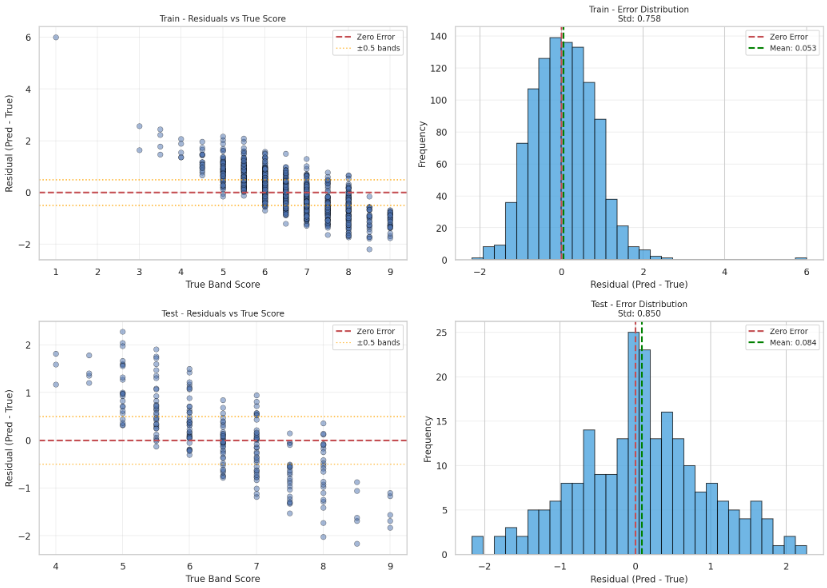}
\caption{(Cycle 4). Confusion matrix of actual vs predicted bands (train/test samples).}
\label{fig:platform}
\end{figure}

Fig. 14 illustrates that the regression models demonstrate stable
predictive performance across training and testing datasets.

In the training set, residuals remain close to zero, indicating accurate
predictions within the 4.0--7.5 band range, with slight deviations in
higher-band essays (7.5--9.0).

The test residuals follow this pattern, showing a slightly wider spread
as expected for unseen data, but a much more balanced error distribution
across the entire IELTS band spectrum.

The model error distribution plots for both train and test data display
typically Gaussian shapes with mean values near zero, confirming that
the model is not systematically overpredicting or underpredicting.

This controlled spread aligns with the MAE scores detailed in Table VIII
and implies that the model generalises well without significant scoring
bias.

Overall, the visual analyses confirm the model's reliability for an
effective adaptive feedback system, providing a solid foundation for
Cycle's adaptive feedback implementation.

\hypertarget{g.-statistical-analysis-of-dbr-cycle-5-results}{%
\subsection{G. Statistical Analysis of DBR Cycle 5 Results}\label{g.-statistical-analysis-of-dbr-cycle-5-results}}

\begin{table}[H]
\centering
\caption{DBR Cycle 5 Pre- and Post-Revision Comparisons}
\begin{tabularx}{\textwidth}{@{}X r@{}}
\toprule
\textbf{Metric} & \textbf{Value} \\
\midrule
\addlinespace[0.3em]

Essay Pairs & 30 \\
\addlinespace[0.3em]

Original Mean $\pm$ SD & $6.724 \pm 0.490$ \\
\addlinespace[0.3em]

Revised Mean $\pm$ SD & $6.784 \pm 0.472$ \\
\addlinespace[0.3em]

Mean Improvement & $+0.06015$ \\
\addlinespace[0.3em]

Essays improved (\%) & 76.7\% \\
\addlinespace[0.3em]

Essays worsened (\%) & 23.3\% \\
\addlinespace[0.3em]

Paired t-test (test statistic, $t$) & 2.714 \\
\addlinespace[0.3em]

Paired t-test (probability, $p$) & 0.01106 \\
\addlinespace[0.3em]

Wilcoxon signed-rank test & $p = 0.00538$ \\
\addlinespace[0.3em]

Cohen's $d$ (paired) & 0.504 (medium) \\
\addlinespace[0.3em]

Significant level ($\alpha$) & 0.05 \\
\addlinespace[0.3em]

\bottomrule
\end{tabularx}
\label{tab:dbr_cycle5_comparison}
\end{table}
The adaptive feedback implementation in DBR Cycle 5 proved to be a
statistically significant positive shift after revision. As indicated by
Table IX, adaptive feedback brought about an +0.060 changes across
personas. Conducting a paired t-test yielded statistics of t = 2.714,
and p = 0.01106 at significance levels ($\alpha$) of 0.05, further validating
that adaptive feedback was significant for the increase in scoring.

Additionally, due to the smaller size sample (30 essays) being used, the
inclusion of the Wilcoxon signed-rank test returning p = 0.00538
excludes that there were any violations of normality in the sample set.

The Cohen's d was included to quantify the standard deviation between
the two grouped means, being the original mean and the revised mean of
the IELTS writing tasks assigned to each persona, reporting a medium
effected value of 0.504 suggests that whilst the change is modest, it is
still meaningfully relative to the subject variability.

The statistical tests conducted on the comparison of the results pre-
and post- adaptive feedback allows us to interpret that the adaptive
feedback system was successful in providing consistent gains across the
sample. Although the improvements were most marginal 0.25-0.5, even
incremental increases in educational setting are particularly relevant,
even more so for higher stake exams such as IELTS Preparations where a
shift of 0.25-0.5 bands can influence outcomes.

\begin{table}[H]
\centering
\small
\caption{DBR Cycle 5 Pre- and Post-Revision IELTS Writing Task Band Scores}
\begin{tabular}{p{2cm} p{1.5cm} p{3cm} p{3cm} p{2.25cm} p{2.25cm}}
\toprule
\textbf{Persona} & \textbf{Essays} & \textbf{Original Mean} & \textbf{Revised Mean} & \textbf{$\Delta$ Mean} & \textbf{\% Improved} \\
\midrule
\addlinespace[0.3em]
P1 & 6 & 6.4630 & 6.5460 & $+0.0830$ & 100\% \\
\addlinespace[0.3em]
P2 & 6 & 6.6849 & 6.7313 & $+0.0463$ & 66.7\% \\
\addlinespace[0.3em]
P3 & 6 & 6.6544 & 6.7254 & $+0.0710$ & 66.7\% \\
\addlinespace[0.3em]
P4 & 6 & 6.9556 & 6.9024 & $-0.0532$ & 50.0\% \\
\addlinespace[0.3em]
P5 & 6 & 6.8619 & 7.0155 & $+0.1536$ & 100\% \\
\addlinespace[0.3em]
\bottomrule
\end{tabular}
\label{tab:dbr_cycle5_personas}
\end{table}

Examining Table X by persona reveals varied effects across feedback
strategies:

\begin{itemize}
\item
  \textbf{P1 (Surface edits)}: Persona 1 saw 100\% essay improvement
  with mean band improvements of (+0.083). Unmasking that cursory
  corrections, primarily grammar, spelling, and punctuation, supplied by
  adaptive feedback consistently resulted in measurable band increases.
\end{itemize}

\begin{itemize}
\item
  \textbf{P2 (Structural) and P3 (Selective)}: While both personas saw
  an increase in mean band score (+0.0463 and +0.0710), changes were
  less uniform in comparison to alternative revision styles, with only
  66.7\% essays improving. This implies that although structural and
  selective modifications were effective, they were not as effective
  across the domain.
\item
  \textbf{P4 (Coherence)}: The coherence-focused revision style yielded
  a small negative average (-0.0532) and only 50\% improved. On the
  exterior this suggests coherence-focused feedback may be riskier
  although a deeper examination could reveal that targeting remediation
  towards one IELTS writing rubric is the real risk. Instead, it would
  have been more beneficial to apply adaptive feedback more
  holistically.
\item
  \textbf{P5 (Conservative)}: Persona 5 capitulated 100\% improved with
  the largest mean increase (+0.1536). Conservative safe edits feedback
  performed to be the most successful at relaying maximal band gains for
  the sample set.
\end{itemize}

Regarding pedagogical implications, the persona outcomes illustrate that
adaptive feedback strategies must balance both risk and reward. Surface
(P1) and conservative (P5) approaches dependably boost scores, whereas
narrower targeted interventions (P2, P3, P4) require cautious
implementations for consistent and positive outcomes.

This observation supports designing adaptive feedback systems that
modulate feedback type based on the student's baseline performance and
multiple revision iterations.

\hypertarget{h.-visual-analysis-of-dbr-cycle-5-results}{%
\subsection{H. Visual Analysis of DBR Cycle 5 Results}\label{h.-visual-analysis-of-dbr-cycle-5-results}}

To complement the statistical tests discussed in Section G, a visual
analysis was performed to verify distributional assumptions,
contextualise effect sizes, and explore how adaptive feedback varies
across different candidate personas.

While the paired t-test and Wilcoxon signed-rank test confirm that the
score improvements are statistically significant, visualisations of
these metrics offer deeper insights into score dispersion, symmetry,
persona-level variability, and the consistency of progress, enabling
statistical inference across the entire sample.

Overall, these figures support the inferential findings by showing that
modest yet systematic score gains arise from different revision
strategies, rather than from outliers or distributional anomalies.

\begin{figure}[H]
\centering
\includegraphics[width=0.9\linewidth,height=8cm]{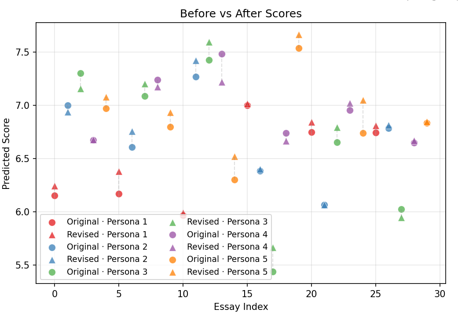}
\caption{(Cycle 5). Before vs After Scores: Scatter Plot (Persona Score Grouping)}
\label{fig:platform}
\end{figure}

Fig. 15's scatter plot shows color-coded personas with essay scores
before (circles) and after (triangles) adaptive feedback revision across
the sample. Most triangle scores are higher than the circle scores,
indicating that adaptive feedback positively affected IELTS writing
extracts and supporting Table VIII\textquotesingle s reported 76.7\%
essay improvement rate.

Notably, Persona 5 (orange) demonstrated the greatest score increase,
highlighting that a conservative revision approach was most beneficial.
While the vertical spread appears tight and the standard deviation
values of ±0.490 and ±0.472 are modest, in the context of a high-stakes
exam like IELTS writing, it is appropriate to consider the increase
statistically significant.

\begin{figure}[H]
\centering
\includegraphics[width=0.9\linewidth,height=8cm]{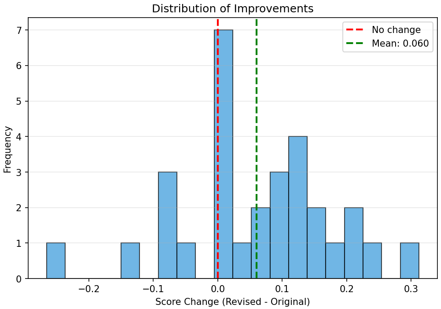}
\caption{(Cycle 5). Histogram Distribution of Essay Improvements}
\label{fig:platform}
\end{figure}

Fig. 16 shows the distribution of band-scores after adaptive feedback
revisions. The red dashed line indicates zero change, while the green
line shows a mean score improvement of 0.060 bands. Most revised essays
lie on the positive side, supporting that 76.7\% improved.

The right-skewed distribution suggests the feedback mainly enhanced
scores, confirmed by a paired t-test with p=0.01106, indicating a 1.1\%
chance results are due to randomness at $\alpha$=0.05.

Since the data aren\textquotesingle t perfectly normal, a Wilcoxon
signed-rank test was used, with p=0.00538, confirming improvements.
Cohen's d of 0.504 indicates a medium effect size, about half a standard
deviation, with small but consistent gains.

The histogram tail extending to +0.30 suggests more substantial
improvements are possible with guided feedback.

\begin{figure}[H]
\centering
\includegraphics[width=0.9\linewidth,height=8cm]{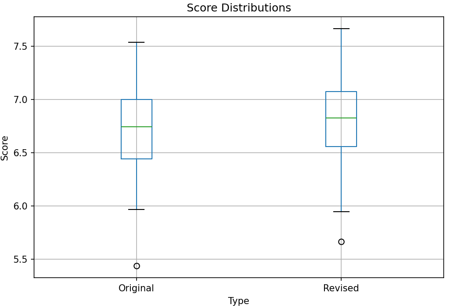}
\caption{(Cycle 5). Box Plot of Score Distributions.}
\label{fig:platform}
\end{figure}

Fig. 17 showcases box plot distributions of essay scoring before and
after adaptive feedback, both containing similar IQR spreads spanning
6.5-7.0 and variability values of 0.47-0.49.

The median of revised scores is slightly higher, reflecting a small mean
improvement (+0.060). Lower outliers remain around 5.5, and the whiskers
and box sizes are nearly identical, indicating that feedback preserves
the score variability.

The overlapping distributions suggest the improvement is modest (Cohen's
d = 0.504). Adaptive feedback raises scores slightly but does not remove
low-performing essays.

Fig. 17 illustrates that adaptive feedback implementation produces
small, consistent gains rather than large jumps in performance.

\begin{figure}[H]
\centering
\includegraphics[width=0.8\linewidth,height=8cm]{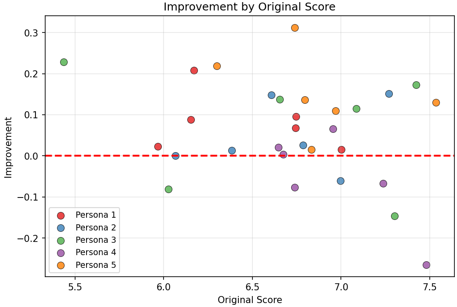}
\caption{(Cycle 5). Scatter plot of Measured Improvement: (Persona Grouping)}
\label{fig:platform}
\end{figure}

Fig. 18 illustrates the extent of improvement, categorised by persona,
clearly visualising the differences in revision styles across personas.
Persona 5 consistently demonstrates positive score gains, whereas
Persona 4 tends to cluster around zero or negative change, especially
for higher-scoring essays.

Overall, Fig. 18 indicates that the level of improvement is influenced
more by the feedback approach than the initial score. This supports the
conclusion that adaptive feedback results in modest but meaningful
improvements, with some strategies proving more effective than others.

\hypertarget{i.-summary-of-dbr-cycles}{%
\subsection{I. Summary of DBR Cycles}\label{i.-summary-of-dbr-cycles}}

The research used a DBR methodology to develop and evaluate an automated
IELTS writing assessment system. Empirical findings from iterative
cycles identified technical and pedagogical limitations, guiding
targeted redesigns to improve grading reliability, generalisation, and
instructional usefulness for IELTS candidates.

Early cycles, especially DBR cycles 2 and 3, established the system
architecture and evaluated rule-based and hybrid grading methods.
Analyses revealed limited accuracy, score compression, and poor
alignment with expert scoring, especially for higher-band essays.
Although incremental improvements were made through refined linguistic
features, these approaches reached a performance ceiling, prompting a
shift away from rule-driven assessment.

DBR Cycle 4 introduced a DistillBERT transformer regression model,
significantly improving evaluation metrics, including reduced MAE,
positive R², and stronger correlations, making it more suitable for
precise IELTS essay grading. Visual analysis confirmed better score
distribution and stability across unseen data, establishing a reliable
scoring backbone for pedagogical interventions.

Building on this, DBR Cycle 5 tested whether adaptive feedback could
support essay improvements in IELTS writing. Using a controlled
framework and a locked test set, statistical testing unveiled a small
but significant improvement in predicted band scores, supported by
visual evidence of consistent gains.

Although modest, these results demonstrate that adaptive feedback can
produce meaningful educational effects in high-stakes assessments when
correctly aligned with learner revision behaviour.

\hypertarget{vii.-discussion}{%
\section{VII. Discussion}\label{vii.-discussion}}

This study investigated how automated scoring combined with adaptive
feedback impacts the IELTS Writing Task 2 utilising DBR cycles to
continuously improve the research quality. Early cycles leverage
rule-based grading systems that were deemed unreliable, producing
inconsistent scores compression around middle bands (5-5.6.5), and
metrics such 1.14 MAE for scoring and negative R² values in Cycles 2 and
3.

Later cycles incorporated a DistilBERT transformer model with addition
of a regression model (Cycle 4), improving overall predictive
performance, delivering a solid predictive base of 0.666 MAE as well as
positively reasoning about essays scoring R² values of 0.495 on the
training set, and 0.354 on the test set demonstrating the model could
generally reason well about data variability.

The solid predictive base permitted the incorporation of adaptive
feedback (Cycle 5), which was tested operating persona revision style
experiments, generating modest but statistically significant score
improvements (mean = +0.060 bands, paired t (29) = 2.714, p = 0.01106
and Wilcoxon p = 0.00538).

Aligned with Ludwig et al. {\cite{ludwig2021automated}}, this research study agrees that
BERT transformer models are better suited for AES, encapsulating and
generalising sensitive linguistic essays more precisely than pure
regressive models, supporting Ludwig et al.'s claim that the transformer
AES implementation correlates greater with human assessors.

Further supporting Ludwig, Cycle 4's transformer model resembled well
with expert-scorers, benchmarking values like Pearson (r) values of
0.60-0.71 and Spearman ($\rho$) of 0.58-0.68 depicting both quality linear
score spacing and ordinal band ranking.

Moreover, this research also supports studies conducted by Ifenthaler
{\cite{ifenthaler2022automated}} reporting stronger performance and more precise score
distribution variability when switching to transformer models with a
regression head in DBR cycle 4 replacing the rule and feature approach
in DBR cycles 2 and 3.

Contrasting platforms more fixated around lower-stakes interactions such
as Duolingo {\cite{shortt2023gamification}} or English advocacy practice such ELSA Speak
{\cite{khoi2024investigating}}. This research emphasises a void on contextualised,
IELTS-specific learning platforms which supply guidance and preparation
for IELTS writing tasks.

The learner persona experimentation underpins that revision behaviour is
a negotiating factor for yielding maximal performance gains through
targeted feedback. Conservative (P1) and cursory (P5) revision
strategies relaying the most consistent and essay score improvements,
with holistic mean improvements of +0.083 and +0.1536.

On the contrary, revision strategies solely focusing on coherence (P4)
occasionally led to score deterioration (-0.0532) and improvements
occurred for only 50\% of essays, implying that revision tactics
neglecting holistic essay structure may be pedagogically
counterproductive.

Results similarly mirrored Bai's {\cite{bai2023application}} controlled classroom study
which yielded high post-test improvement in lexical resource and grammar
accuracy following writing guidance led by ChatGPT, but only slight
inconsistent improvements occurred when following coherence guidance.
This research backs Bai's claim that higher-order cognitive writing
demands are required for a deeper structuring rather than surface level
refinements such as lexical resource and grammatical corrections.

Pedagogically adaptive feedback systems in critical writing scenarios
should focus on safety, explainability, and incremental remediation, as
shown by conservative grammatical and lexical corrections, while
aggressive restructuring can sometimes negate potential improvements
{\cite{bai2023application}}. Additionally, results favoured a hybrid model where automated
feedback is supplementary to instructor guidance rather than replacing
it entirely.

Although this study was measuring the influence of adaptive feedback was
controlled through persona-based case studies, system design remains
crucial for smoothly integrating adaptive feedback into a learning
platform, especially for ESL learners when tailoring for exams like the
IELTS. Aligning with cognitive load theory and prior UI research that
emphasise structured feedback hierarchies, avoiding divided attention
and task ambiguity {\cite{fang2025effects}}.

Research limitations include the reliance on a small test set of 30
writing samples and simulated persona revisions, opposed to real IELTS
candidates, limiting potential insights into long-term pedagogical
effects. Although the DistilBERT model achieved good precision,
underpredictions lingered between higher band ranges, particularly the
8.0-9.0 bands.

Despite better accuracy than Cycles 2 and 3 rule-based methodologies,
persistent underpredictions could affect the platform\textquotesingle s
usefulness for more advanced writers. However, since the large majority
of IELTS candidates are more focused on passing rather the examination
than achieving perfect scores, deeming underprediction to be a less
critical limitation.

Finally, system\textquotesingle s diagnostic reports statistically and
visually, support pedagogical unbiased effectiveness and writing
improvements, validation from official IELTS examiners would be
essential for future system refinements.

Future research should mainly focus on feedback from extensive real
IELTS candidate studies, tracking essay improvements across multiple
longer-term revision cycles, assessing score changes and perceived
platform usefulness.

Subsequently, experiments that break down individual IELTS feedback
writing components such as grammar, lexical resource, coherence, and
task achievement on a sufficiently large scale could reveal which
factors most significantly enhanced IELTS scores. They can also identify
which rubric elements candidates emphasise most, highlighting the key
areas needing immediate intervention.

When conducting these user studies, it is crucial to consider potential
biases related to writing standards, especially due to heavy reliance on
AI technologies. Counteractively implementing anti-detection measures,
like AI detection, is essential to eliminate non-authentic examples and
preserve the validity of the case study {\cite{moran2023chatgpt}}.

\section{VIII. Conclusion}
\label{sec:conclusion}

Above, we analysed and discussed the benefits of curating a IELTS
writing task learning platform featuring AES in conjunction with
pedagogically aligned adaptive feedback through utilising DBR cycles to
progress automated essay scoring from rule-based systems to a
transformer-based model with a regression head.

Early cycles identified the limitations of feature and rule-based
methods, while integrating a DistilBERT transformer model crafted a more
stable predictive foundation. Adding adaptive feedback to this setup
showed that even modest, statistically significant essay improvements
are possible when feedback is based on consistent automated assessment.

Notably, persona-based adaptive feedback revealed that learner revision
behaviour influences feedback success. Conservative grammatical and
lexical guidance delivered the most consistent gains, while narrowly
focused coherence interventions were less reliable, highlighting the
risks of aggressive essay restructuring {\cite{bai2023application}}.

The findings suggest that adaptive feedback should complement human
instruction rather than replace it, especially for rubric elements like
Task Achievement and Coherence and Cohesion that need restructuring or
deeper analysis, which would benefit from guidance and interpretation
that align with pedagogical goals. Conversely, elements such as
Grammatical Range and Accuracy, and Lexical Resource can be addressed
effectively with minimal human intervention.

Future research could explore embedding real ESL candidate evaluations
and human-in-the-loop validation to assess long-term learning transfer
and fairness in assessment contexts.

\pagebreak

\bibliographystyle{IEEEtran}
\bibliography{references}

\end{document}